\documentclass[nohyperref]{article}

\usepackage{hyperref}

\usepackage[accepted]{icml2023}

\usepackage{amsmath}
\usepackage{amssymb}
\usepackage{mathtools}
\usepackage{amsthm}
\usepackage[capitalize,noabbrev]{cleveref}
\usepackage[utf8]{inputenc} % allow utf-8 input
\usepackage[T1]{fontenc}    % use 8-bit T1 fonts
\usepackage{url}            % simple URL typesetting
\usepackage{booktabs}       % professional-quality tables
\usepackage{amsfonts}       % blackboard math symbols
\usepackage{nicefrac}       % compact symbols for 1/2, etc.
\usepackage{microtype}      % microtypography
\usepackage{graphicx}
\usepackage{wrapfig}
\usepackage{mathtools}
\usepackage{amsthm}
\usepackage{enumitem}
\usepackage{subcaption}
\usepackage{algpseudocode}
\usepackage{multirow}
\usepackage[justification=centering]{caption}

\usepackage{pifont}

\usepackage[capitalize,noabbrev]{cleveref}

\theoremstyle{plain}
\newtheorem{theorem}{Theorem}[section]

\theoremstyle{definition}

\theoremstyle{remark}
\newtheorem{remark}[theorem]{\textbf{Remark}}

\usepackage[textsize=tiny]{todonotes}

\definecolor{forestgreen}{RGB}{34,159,44}
\definecolor{deepred}{RGB}{235,0,20} %{256,50,100}

\newcommand*{\green}{\textcolor{forestgreen}}
\newcommand*{\deepred}{\textcolor{deepred}}

\usepackage{amssymb}% http://ctan.org/pkg/amssymb
\usepackage{pifont}% http://ctan.org/pkg/pifont
\newcommand{\cmark}{\ding{51}}%
\newcommand{\xmark}{\ding{55}}%

\newcommand{\mycmark}{\green{\cmark}}
\newcommand{\myxmark}{\deepred{\xmark}}

\newcommand{\bx}{\mathbf{x}}

\newcommand{\bz}{\mathbf{z}}
\newcommand{\by}{\mathbf{y}}
\newcommand{\btheta}{\mathbf{\theta}}

\newcommand{\cH}{\mathcal{H}}

\newcommand{\cN}{\mathcal{N}}

\newcommand{\bE}{\mathbb{E}}

\icmltitlerunning{Restoration based Generative Models}

\begin{document}
\twocolumn[
\icmltitle{Restoration based Generative Models}




% \begin{icmlauthorlist}
% \icmlauthor{Jaemoo Choi}{equal}
% \icmlauthor{Yesom Park}{equal}
% \icmlauthor{Myungjoo Kang}{}
% \end{icmlauthorlist}
% \vspace{5pt}
% % \let\thefootnote\relax\footnotetext{\textsuperscript{*}Equal contribution authors. Correspondence to: \tt <mkang@snu.ac.kr>.}
% \Info

% % \icmlcorrespondingauthor{Firstname1 Lastname1}{first1.last1@xxx.edu}
% % \icmlcorrespondingauthor{Firstname2 Lastname2}{first2.last2@www.uk}

% % You may provide any keywords that you
% % find helpful for describing your paper; these are used to populate
% % the "keywords" metadata in the PDF but will not be shown in the document

% \vskip 0.3in
% ]
% \msgs

\icmlsetsymbol{equal}{*}

\begin{icmlauthorlist}
\icmlauthor{Jaemoo Choi}{equal,snu}
\icmlauthor{Yesom Park}{equal,snu}
\icmlauthor{Myungjoo Kang}{snu}
\end{icmlauthorlist}

\icmlaffiliation{snu}{Department of Mathematical Sciences, Seoul National University, Seoul, South Korea}

\icmlcorrespondingauthor{Myungjoo Kang}{mkang@snu.ac.kr}

\icmlkeywords{Generative models, Denoising diffusion models, ICML}

\vskip 0.3in
]



\printAffiliationsAndNotice{\icmlEqualContribution} % otherwise use the standard text.

\begin{abstract}
Denoising diffusion models (DDMs) have recently attracted increasing attention by showing impressive synthesis quality. DDMs are built on a diffusion process that pushes data to the noise distribution and the models learn to denoise. 
In this paper, we establish the interpretation of DDMs in terms of image restoration (IR). Integrating IR literature allows us to use an alternative objective and diverse forward processes, not confining to the diffusion process. By imposing prior knowledge on the loss function grounded on MAP-based estimation, we eliminate the need for the expensive sampling of DDMs. Also, we propose a multi-scale training, which improves the performance compared to the diffusion process, by taking advantage of the flexibility of the forward process. Experimental results demonstrate that our model improves the quality and efficiency of both training and inference.
Furthermore, we show the applicability of our model to inverse problems. We believe that our framework paves the way for designing a new type of flexible general generative model.
% \red{\textbf{git code}}
\end{abstract}

\section{Introduction}
\begin{figure}[t]
  \begin{center}
    \includegraphics[width=0.45\textwidth]{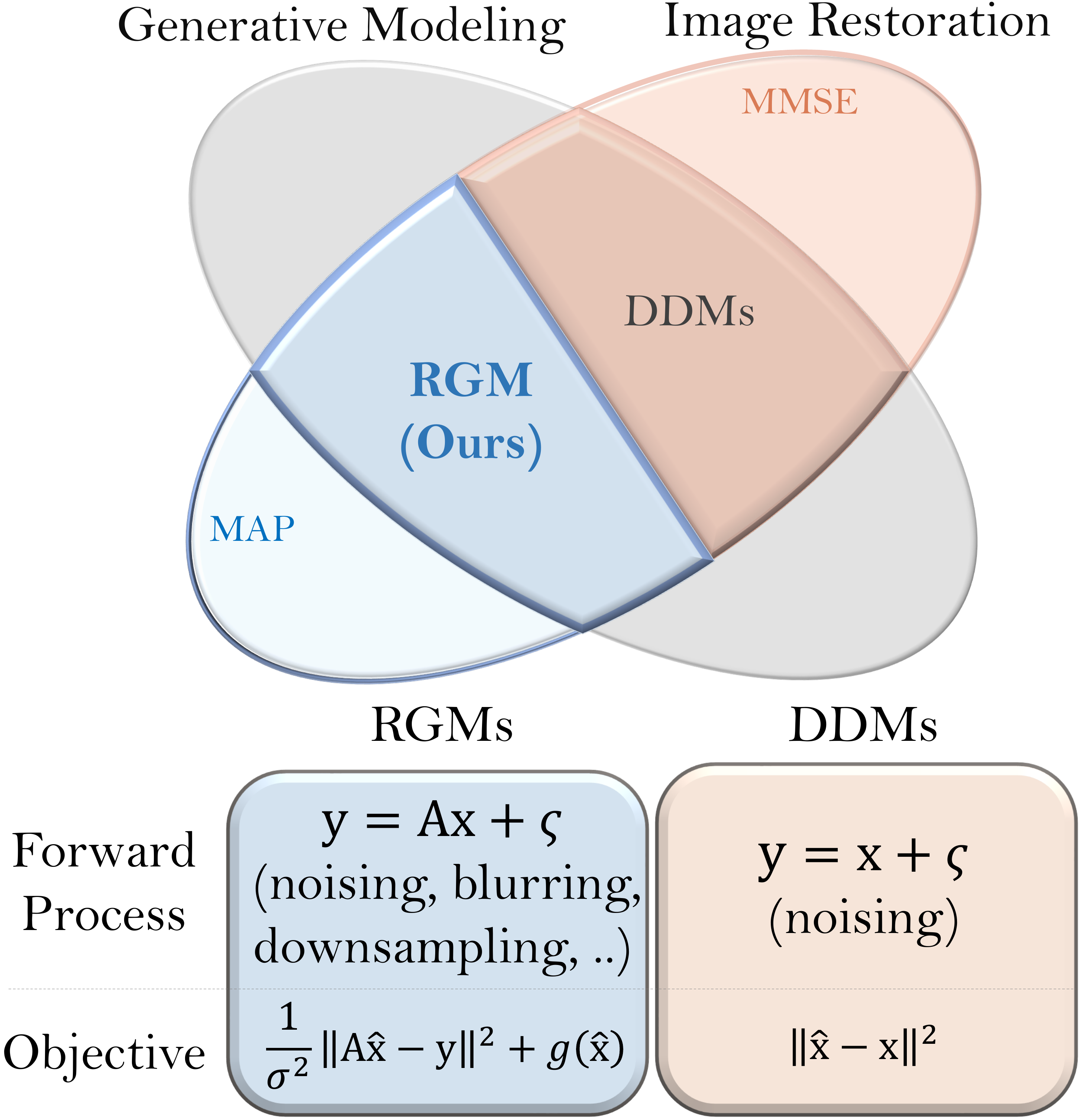}
  \end{center}
  \vspace{-10pt}
  \caption{Conceptual comparison of RGMs and DDMs. Here, $\bx$ is the original data, $\by$ is the degradation of $\bx$ with noise $\xi \sim \mathcal{N}(\mathbf{0} , \sigma^2 \mathbf{I})$, and $\hat{\bx}$ is the reconstruction of $\by$.}
  \vspace{-12pt}
  \label{fig:main}
\end{figure}
Generative modeling is a prolific machine learning task that the models learn to describe how a dataset is distributed and generate new samples from the distribution.
The most widely used generative models primarily differ in their choice of bridging the data distribution to a tractable latent distribution \cite{goodfellow2020generative, kingma2013auto, rezende2014stochastic, rezende2015variational, sohl2015deep, chen2021likelihood}.
In recent years, denoising diffusion models (DDMs) \cite{ddpm, song2019generative, scoresde, dockhorn2021score} have drawn considerable attention by demonstrating remarkable results.
%in terms of both high sample quality and likelihood.
DDMs rely on a forward diffusion process that progressively transforms the data into Gaussian noise, and they
learn to reverse the noising process.
Albeit their enormous successes, their gradual denoising generative process gives rise to low inference efficiency. To pull latent variables back to the data distribution, the denoising process often requires hundreds or even thousands of network evaluations to sample a single instance.
% In addition, as DDMs couple the data and latent variables through the diffusion process, the latent and data possess the same dimension. %is of the same dimension as the data.
Many follow-up studies consider enhancing inference speed \cite{ddim, tachibana2021taylor, lu2022dpm} or grafting with other generative models \cite{xiao2021tackling, vahdat2021score, zhang2021diffusion, pandey2022diffusevae}.
% However, most of these works still require numerous steps to generate high-fidelity samples and face latent inefficacy.

In this study, we focus on a different perspective. We interpret the DDMs through the lens of image restoration (IR), which is a family of inverse problems for recovering the original images from corrupted ones \cite{castleman1996digital, gunturk2018image}. The corruption arises in various forms, including noising \cite{rudin1992nonlinear,buades2005non} and downsampling \cite{farsiu2004fast}. IR has been a long-standing problem because of its high practical value in various applications \cite{besag1991bayesian, banham1997digital, lehtinen2018noise2noise, ma2011low}. From the IR point of view, DDMs can be considered as IR models based on minimum mean square error (MMSE) estimation \cite{zervakis1991class, pnpuld}, focusing only on the denoising task. Mathematically, IR is an ill-posed inverse problem in the sense that it does not admit a unique solution \cite{hadamard1902problemes} and hence, leads to instability in reconstruction. Owing to the ill-posedness of IR, MMSE produces impertinent results. DDMs alleviate this problem by leveraging costly stochastic sampling, which has been regarded as an indispensable tool in the literature on DDMs.  
% By casting DDMs as IR models, however, the forward process need not be restricted to Gaussian noising, and ill-posedness can be detoured in ways other than costly iterative sampling.

Inspired by this observation, we propose a new flexible family of generative models that we refer to \textit{restoration-based generative models (RGMs)}. 
We adopt an alternative objective; a maximum a posteriori (MAP) based regularization \cite{hunt1977bayesian,trussell1980relationship}, which is predominantly used in IR.
%First, we adopt an alternative objective; a maximum a posteriori (MAP) \cite{trussell1980relationship, hunt1977bayesian}, which is predominantly used in IR.
This approach detours the ill-posedness by regularizing the data fidelity loss by prior knowledge rather than doing costly iterative sampling.
Prior knowledge can be utilized in a variety of ways. 
Moreover, we also have the freedom to design the degradation process. RGMs with variability of these two have several benefits:
\vspace{-7pt}
\paragraph{Implicit Prior knowledge}
% Prior knowledge should be carefully designed because it affects the quality of the restorations and ill-posedness mitigation.
Many hand-crafted regularization schemes \cite{tikhonov1963regularization, donoho1995noising, baraniuk2007compressive} encourage solutions to satisfy certain properties, such as smoothness and sparsity.
However, for the purpose of density estimation, we parametrize the prior term to learn the statistical distance, e.g., Kullback–Leibler divergence or Wasserstein distance.
% However, for the purpose of density estimation, we implicitly parameterize the prior term by leveraging existing provable generative models, such as GAN discriminator \cite{goodfellow2020generative}.
We also introduce a random auxiliary variable to further ease the ill-posedness. 
Our MAP-based estimation allows a much smaller computational cost, retaining the density estimating capability of DDMs.
% \textcolor{red}{Our regularization method retains the density estimating capability of DDMs at a much smaller computation cost.}
% Our MAP approach retains the density estimating capability of DDMs at a much smaller computation cost.
% Our MAP objective frees us from the costly sampling of DDMs. 
\vspace{-7pt}
\paragraph{Various Forward process} DDMs are buried in a Gaussian noising process. On the other hand, fluidity in the forward process of RGMs improves model performance because the behavior of generative models is significantly affected by how the data distribution is transformed into a simple distribution.
As one instantiation, we design a degradation process that progressively reduces the dimension by block averaging the image, which improves performance.
% On the other hand, because the behavior of generative models is significantly affected by how the data distribution is transformed into a simple distribution, fluidity in the forward process of RGMs permeate benefits from it.
% Unlike DDMs, which are buried in a Gaussian noising process, RGMs can be combined with other general degradation processes.
% As one instantiation, we design a multi-scale training that resolves the latent inefficiency of DDMs.

Our comprehensive empirical studies on image generation and inverse problems demonstrate that RGMs generate samples rivaling the quality of DDMs with several orders of magnitude faster inference. In particular, our model achieves FID 2.47 on CIFAR10, with only seven network function evaluations. 
Furthermore, through rigorous experiments with various prior terms and degradation, we validate that our RGM framework is well-structured that opens the way for designing more efficient and flexible generative models.
\vspace{-7pt}
\section{Background} \label{sec:background}
\paragraph{Image Restoration}
A common inverse problem arising in image processing, including denoising and inpainting, is the estimation of an image $\bx$ given a corrupted image 
\begin{equation} \label{eq:inverse}
\mathbf{y}=\mathbf{A}\mathbf{x} + \mathbf{\xi},
\end{equation}
where $\mathbf{A}$ is a matrix that models the degradation process, and $\xi \sim \cN\left(0,\mathbf{\Sigma}\right)$ is an additive noise.
A family of such problems is known as image restoration (IR).
% A commonly used Bayesian approach 
A popular approach is the maximum-a-posteriori (MAP) estimation
\begin{equation} \label{eq:baysian_map}
\mathrm{argmax}_{\mathbf{x}}\ \log p\left(\mathbf{x}\mid\mathbf{y}\right)
= \log p\left(\mathbf{y}\mid\mathbf{x}\right) + \log p\left(\mathbf{x}\right).
\end{equation}
However, since the explicit density function $\log p\left(\mathbf{x}\right)$ is intractable, they replace the objective \eqref{eq:baysian_map} with
\begin{equation}\label{eq:map}
\mathrm{argmin}_{\mathbf{x}}\ f\left(\mathbf{y},\mathbf{x}\right) + g\left(\mathbf{x}\right),
\end{equation}
where $f\left(\mathbf{x},\mathbf{y}\right)$$=$$-$$\log p\left(\mathbf{y}\mid \mathbf{x}\right)$$=\frac{1}{2}\left\Vert \left(\mathbf{\Sigma}^{\dagger}\right)^{\frac{1}{2}}\left(\mathbf{A}\mathbf{x}-\mathbf{y}\right)\right\Vert ^{2}_2$ is the data fidelity term with the pseudoinverse \cite{moore1920reciprocal} $\mathbf{\Sigma}^{\dagger}$ and $g$ is a regularization term (or prior knowledge) that represents prior or constraints on the solution.
Since \eqref{eq:map} originated from the MAP objective, it is also called the MAP-based approach.
% Prior knowledge can be imposed in a variety of ways.
% The choice of the prior $g$ is crucial because the quality of the restoration varies according to different prior.
% Priors can be adapted depending on various tasks
Prior knowledge can be imposed in a variety of ways and the choice is crucial because the quality of the restoration varies according to different prior.
Moreover, the ill-posedness nature of the inverse problem \eqref{eq:inverse}, that is non-uniqueness of the solution, necessitates the use of regularization.
By imposing certain prior information about the desirable recovery, the prior knowledge $g$ relieves the ill-posedness.
% Moreover, the $g$ relieves the ill-posedness nature of the inverse problem \eqref{eq:inverse}, that is non-uniqueness of the solution, by imposing the assumption about the desirable solution.
% These necessitate the use of regularization.
Therefore, many researchers have been devoted to designing a proper prior $g$  \cite{rudin1992nonlinear, mallat1999wavelet, lunz2018adversarial}.

\paragraph{Denoising Generative Models}
Denoising diffusion models (DDMs) \cite{ddpm, scoresde}
%, such as DDPM \cite{ddpm}, and score matching with Langevin dynamics \cite{scoresde}, 
have recently emerged as the forefront of image synthesis research. 
Starting from the image distribution, they gradually corrupt the image $\bx_0\sim p_{\text{data}}$ into Gaussian noise over time through a forward diffusion process with a given noise schedule $\sigma_t$:
\[
q^{(t)}\left(\bx_{t}\mid\bx_{0}\right)=\cN\left(\bx_0,\sigma_t^2\mathbf{I}\right),
\]
also known as VESDE \cite{scoresde}. Another linear diffusion process named VPSDE is also leveraged, but since these two are known to be exchangeable with each other \cite{kim2022soft}, the paper focuses on VESDE.
DDMs pose the data generation as an iterative denoising procedure by learning the reverse of the forward process.
As they are modeled with conditional Gaussian distributions, evidence lower bound (ELBO) \cite{sohl2015deep} could be simplified to the following objective  with a weight $\lambda\left(t\right)\geq0$ \cite{ddpm, scoresde}:
\begin{equation} \label{eq:scorematching}
\Sigma_{t=0}^T\bE_{\bx_0 \sim p_\text{data},\bx_{t}\sim q_{\sigma_t}\left(\bx_t\mid\bx_0\right)}\left[\lambda\left(t\right)\lVert G_\theta\left(\bx_t,t\right)-\bx_0\rVert ^2_2\right],
\end{equation}
where $G_\theta$ is a neural network parametrized by $\theta$.

For each forward step $t$, \eqref{eq:scorematching} is the minimum mean square error (MMSE) objective.
MMSE loss is simple and straightforward to train, however, the solution is only optimized to ensure accordance with the degradation process because it only contains the recovery term. Consequently, MMSE is affected by ill-posedness.
To be precise, when $\sigma_t$ is large, \eqref{eq:inverse} becomes highly ill-posed and possesses many solutions for a given observation.
In this case, the MMSE solution averages all these candidate solutions, resulting in an atypical reconstruction. 
Recent works \cite{pnpuld, kawar2021snips} have endeavored to resolve this problem by stochastic sampling, however, they suffer from notoriously low efficiency as they roll out thousands of trajectories. 
In a similar manner, DDMs utilize a sampling scheme that often requires hundreds to thousands of steps.
In summary, there are two limitations of DDMs from the IR perspective:
\begin{enumerate}[leftmargin=10pt, topsep=0pt]
\item The degradation process is restricted to Gaussian noising.
\vspace{-15pt}
\item The inference efficiency is intrinsically low due to the MMSE estimator.
\vspace{-0pt}
\end{enumerate}

\section{Method} \label{section:method}

\subsection{MAP-based Estimation for Generation} \label{sec:RGM}
As alluded in \cref{sec:background}, DDMs can be regarded as MMSE-grounded IR models, specialized in denoising.
This observation brings us a new perspective on the design of a family of flexible generative models.
As an alternative to MMSE, we propose a new generative model based on \eqref{eq:map}:
\begin{equation} \label{eq:map_gen}
    \mathbb{E}_{\bx \sim p_\text{data}, \by \sim \cN\left(\bx, \sigma^2 \mathbf{I}\right)}\left[\frac{1}{2\sigma^2} \lVert G_\btheta (\by) - \by \rVert_2^2  +\lambda g(G_\theta(\by)))\right],
\end{equation}
where the first term measures the data fidelity and the second term delivers the prior knowledge of the data distribution.
It has been adopted as a standard approach for high-dimensional imaging problems and is known to be more relevant than MMSE in many applications \cite{saha2009particle, bigdeli2019image,chen2016higher}.
% MAP can sidestep the time-consuming iterative inference of MMSE
By leveraging prior information on the solution, MAP-based approaches alleviate the ill-posedness of the inverse problem \eqref{eq:inverse}, without the use of costly sampling.
% By leveraging prior information on the solution, MAP-based approaches alleviate the ill-posedness of the inverse problem \eqref{eq:inverse}, without the use of costly sampling.
Therefore, carefully crafting the relevant prior term is crucial. We now show how one can execute an appropriate prior term for density estimation while alleviating the ill-posedness.

\paragraph{Alleviation of ill-posedness}
Unlike the generic denoising task, it is necessary to bridge the image to the Gaussian noise to learn the data distribution. As the noise level increases, a single distorted observation has several solutions, which indicates that the ill-posedness deepens. Therefore, our generator $G_\theta$ should  be able to recover diverse restorations from a degraded image to express the data distribution more abundantly.
Since it is difficult for the regularization term to remedy all these problems on its own, we further offload the ill-posedness by introducing a random auxiliary variable $\bz\sim\cN\left(\bz\mid \mathbf{0},\mathbf{I}\right)$.
In other words, we use the stochastic variable $\bz$ as an input of $G_\theta$. As $G_\theta\left(\by,\bz\right)$ generates various restores for different $\bz$, it is more amenable to faithfully recovering the data distribution.
% In other words, we use not only a degraded image $\by$ but also the stochastic variable $\bz$ as inputs of $G_\theta$.

\paragraph{Implicit Prior Knowledge} 
For density estimation, the knowledge about the data distribution should be properly encoded in the prior term $g$ of \eqref{eq:map_gen}.
% \textcolor{red}{However, since the explicit density of the data is inaccessible, we design $g$ to learn the prior knowledge.}
% \textcolor{red}{The explicit density function with determined marginal distribution is intractable, but it can be learned by parametrizing the prior term.}
However, since the explicit density of the data is inaccessible, we parameterize $g$ to learn the prior.
%with the aid of well-developed generative models.
% In the Bayesian estimation there is a wide freedom in the choice of prior. 
For each forward step, our new objective for the generator $G_\theta$ in conjunction with the $\bz$ is given by:
\begin{align}
\begin{split}
   \mathbb{E}_{\bx \sim p_\text{data}, \by \sim \cN\left(\bx, \sigma^2 \mathbf{I}\right), \bz \sim \mathcal{N}(0, I)} &  \left[\frac{1}{2\sigma^2} \lVert G_\theta(\by, \bz) - \by  \rVert_2^2 \right. \\ 
   &\ \ \  + \left.\lambda g_{\phi}(G_\theta(\by, \bz)))\right],
\label{eq:RGM_Dloss}
\end{split}
\end{align}
where $g_\phi$ is a learnable prior term parameterized by $\phi$.
%, which is designed by integrating various existing generative models, and $\lambda \geq 0$ is a hyperparameter.
\begin{figure*}[t]
  % \vspace{-5pt}
  \begin{center}
    \includegraphics[width=0.45\textwidth]{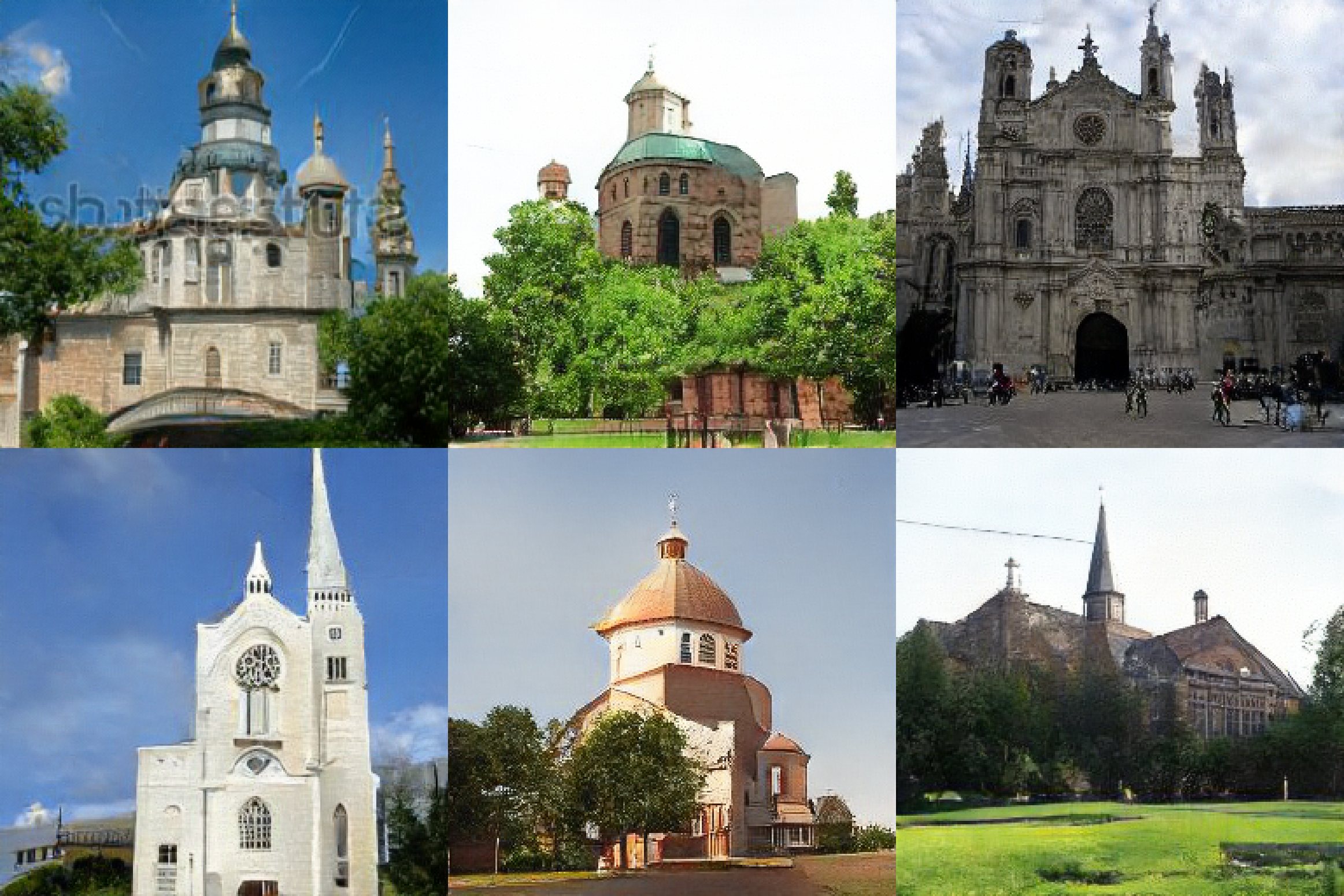}
    \hfill
     \includegraphics[width=0.45\textwidth]{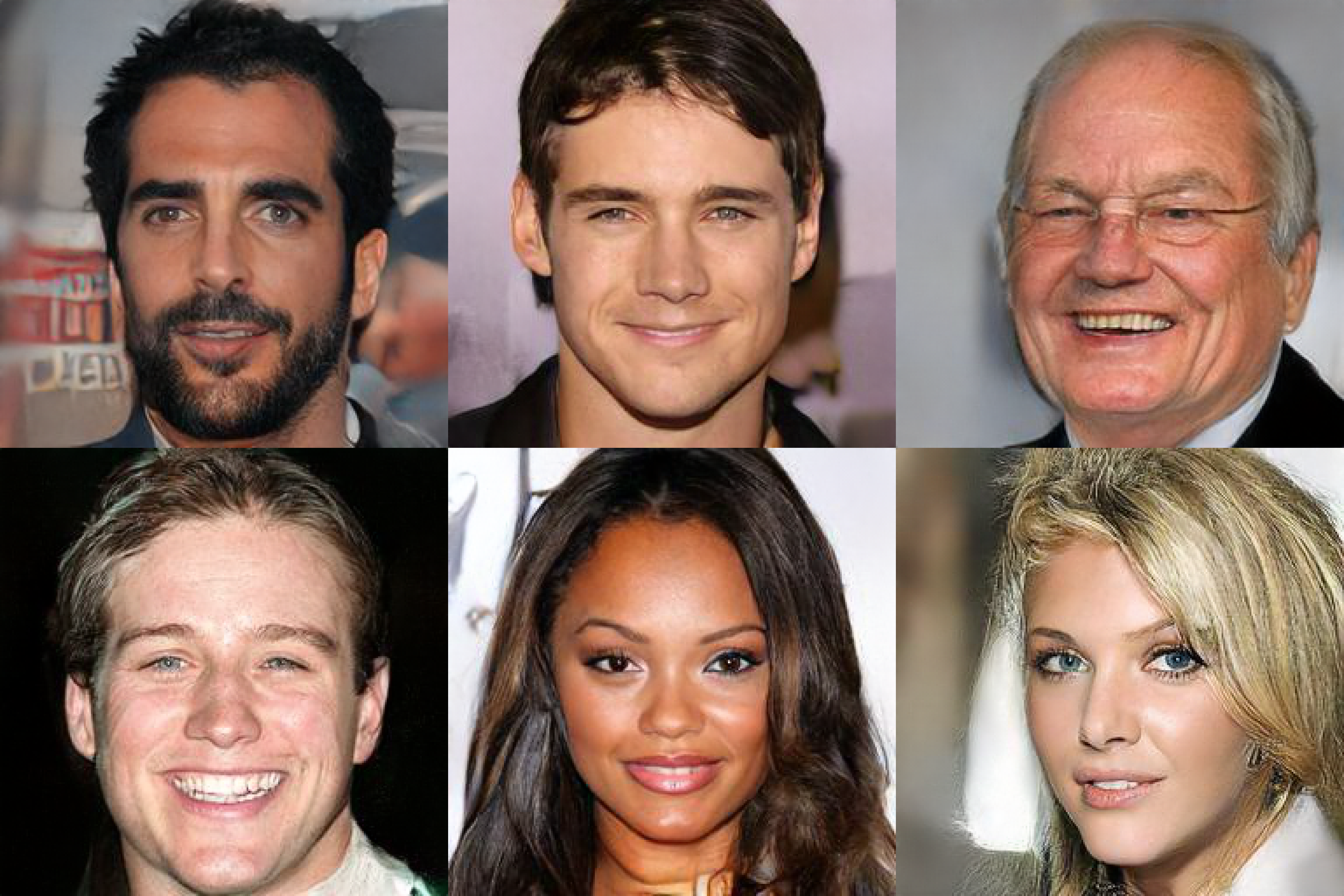}
  \end{center}
  \vspace{-10pt}
  \caption{Generated samples on LSUN Church (left) and CelebA-HQ (right).} \label{fig:samples}
    \vspace{-15pt}
\end{figure*}
For example, we can learn an implicit representation of the data by adopting $g_{\phi}(\bx)=\log\left(1 - D_\phi \left(\bx\right)\right) - \log D_\phi \left(\bx\right)$ where $D_\phi$ is a discriminator trained coupled with $G_\theta$.
As $D_\phi$ gets close to the optimal, i.e., $D_\phi(x) = \frac{p_{\text{data}}(x)} {p_{\text{data}}(x)+p_\theta(x)}$, the expectation of $g_\phi$ approaches to Kullback–Leibler divergence (KLD) $D_{\text{KL}}(p_\theta||p_{\text{data}})$, which leads the loss \eqref{eq:RGM_Dloss} with $\lambda=1$ to agree with the following objective:
% $D_{KL}(p_\theta||p_{\text{data}})$
% adopting discriminator $D_\phi$
%For example, we can learn an implicit representation of the data density by adopting a discriminator generative adversarial network (GAN), 
%which has shown promising results in many generative tasks, 
%and use generator loss as a relevant prior term
% $g_{\phi}(\bx)=\log\left(1 - D_\phi \left(\bx\right)\right) - \log D_\phi \left(\bx\right)$ that is trained coupled with a discriminator $D_\phi$.
% In our experiments, $D_\phi$ and $g_\phi$ tries to maximize, minimize Jensen-Shannon divergence (KLD), respectively.
%  Note that when the discriminator is optimal, i.e. $D_\phi(x) = \frac{p_{\text{data}}(x)}{p_{\text{data}}(x)+q_\theta(x)}$, the loss \eqref{eq:RGM_Dloss} with $\lambda =1$ agrees with the following objective: %to learn the posterior distribution
\begin{equation}
   \mathbb{E}_{\bx \sim p_\text{data}, \by \sim \cN\left(\bx, \sigma^2 \mathbf{I}\right)}\left[ D_{KL}(p_\theta (\bx|\by)|| p(\bx|\by)) \right]+\mathcal{H}(p_\theta),
\end{equation}
where $\cH$ denotes an entropy and $p_\theta$ the distribution generated by $G_\theta$.
%, which corresponds to training the model to learn the posterior distribution.
The overall training procedure combined with all $\sigma\in \{\sigma_k\}_{k=1}^T$ is provided in \cref{appen:train}.
\begin{remark}
Contrary to conventional IR literature whose prior term is pre-defined, our approach \eqref{eq:RGM_Dloss} tries to learn the prior term by coordinating with $G_\theta$.
%through the discriminator.
This end-to-end training allows our 
MAP-inspired scheme to deliver more promising performance.
Moreover, it is worth noting that 
our framework has the wide freedom in the choice of $g_\phi$
%by any other generative model, 
without being tied to the discriminative learning for KLD exemplified above.
% we can not only parameterize the prior term by GAN, but also by any generative model. 
Consequently, we further design the prior term by mounting maximum mean discrepancy (MMD) \citep{dziugaite2015training} and distributed sliced Wasserstein distance (DSWD) \citep{nguyen2020distributional} in \cref{sec:experiments}.
\end{remark}
% As conventional, we train the discriminator $D_\phi$ to minimize the Jensen-Shannon divergence, which approximates $D_\phi\left(\bx\right) \approx p_\text{data}\left(\bx\right) / \left(p_\text{data}\left(\bx\right)+q_\theta \left(\bx\right)\right)$ where $q_\theta$ is the generator distribution.
\vspace{-5pt}
\paragraph{Small Denoising Steps}
A major downside of DDMs is their sampling inefficiency, 
which often requires hundreds to thousands of denoising steps to obtain a single image. 
%Therefore, there have been several efforts to improve the sampling speed of \red{DDM}s \cite{ddim, ddgan}.
By adopting regularization term $g$,
%By adopting MAP approach and parametrizing the prior distribution through generative models, 
our approach provides an avenue to offload the time-consuming sampling and enables significantly small denoising steps.
For small degradation, we can obtain a restored image in one shot. But, as our restoration starts from the Gaussian noise, the data distribution is not completely estimated. Therefore, we perform the generation iteratively.
In our experiments on CIFAR10, we generate a high-quality sample in four denoising steps.

\subsection{Extension to General Restoration} \label{sec:forward}
In Section \ref{sec:RGM}, we proposed a denoising generative model based on MAP objective. However, from the IR perspective, it is not necessary to restrict the forward process to Gaussian noising ($\mathbf{A}=\mathbf{I}$) and it can be generalized to any family of degradation matrices $\mathbf{A}$ and noise factors $\mathbf{\Sigma}$ in \eqref{eq:inverse}.
% by replacing the forward process and the loss function.
% We consider a general forward process in which the image is distorted by a linear operation $\mathbf{A}$ and contaminated by noise $\xi \sim \cN\left(0,\sigma^2 \mathbf{I}\right)$. 
Utilizing the general forward process, we can learn the generative model
by generalizing the loss function \eqref{eq:RGM_Dloss} as follows:
\begin{align} \label{eq:RGMloss}
\begin{split}
    & \bE_{\bx\sim p_\text{data}, \by \sim \cN(\mathbf{A}\bx, \mathbf{\Sigma}), \bz \sim \cN(\mathbf{0}, \mathbf{I})}\left[\lambda g_{\phi}(G_\theta(\by, \bz))) \right.\\
    &\ \ \ \ \ \ \ \ \ \ \ \ \ \ \ \ \ \  \left. + \frac{1}{2} \left\Vert \left(\mathbf{\Sigma}^{\dagger}\right)^{\frac{1}{2}}\left(\mathbf{A}\cdot G_\theta(\by, \bz) - \by \right) \right\Vert_2^2 \right].
    \end{split}
\end{align}
Therefore, RGM has an flexible structure that can permeate any forward process, and aids in designing a new generative model.
%allowing us to permeate benefits from the IR literature, and aids in designing a new generative model.
% It guides us in designing a new generative model. 
Here, we propose a new model established upon super-resolution (SR).
\vspace{-5pt}
\paragraph{Multi-scale RGM}
Most DDMs maintain the image size during the diffusion process by adding noise to individual pixels. Consequently, they are very inefficient because they require a latent as much as dimension of pixel space that is much larger than the submanifold of the image space.
Motivated by this, we take 
$\mathbf{A}$ as a block averaging filter that averages out $2\times 2$ pixel values.
Halving the image size at each coarsening step allows us a more expressive generative model with a lower-dimensional latent distribution.
Moreover, multi-scale training has proven to be an effective strategy
for synthesizing large scale images \cite{denton2015deep, progan, reed2017parallel}.
Therefore, our model produces strikingly realistic images by progressively extracting spatial information.
%in coarse resolution.    

\section{Experiments} \label{sec:experiments}

This section evaluates the performance of the proposed RGMs on synthetic and several benchmark datasets.
% , including CIFAR10 \cite{krizhevsky2009learning} ($32\times 32$ unconditional), CelebA-HQ \cite{liu2015deep} ($256\times 256$), and LSUN Church \cite{yu2015lsun} ($256\times 256$). 
We also analyze our model through extensive ablation studies.
Furthermore, we show the capability of RGMs for solving inverse problems. 
We parametrize $G_\theta$ based on the UNet-like structure \cite{ronneberger2015u} which was successfully used in NCSN++ \cite{xiao2021tackling}.
The internal details of the implementation can be found in Appendix \ref{appen:implement_detail}. 

\paragraph {Setup}
% We implement two models: \textit{RGM-KLD-D} is a model trained with the diffusion process, which is mainly used by DDMs. We also consider a multi-scale model whose degradation matrix is a $2\times 2$ averaging filter.
% In this case, unlike the diffusion process, the image is corrupted by a downsampling filter together with additive noise. Therefore, the model (termed by \textit{RGM-KLD-SR (naive)}) is demanded to conduct upsampling and denoising at the same time. 
% To make it more effective, we explore another forward schedule that separates the downsampling and noising process and performs them alternatively. \textit{RGM-KLD-SR} refers to the model to which this schedule is applied. (See Appendix \ref{appen:shcedule} for details). 

Our RGMs have a free hand in designing the forward process (i.e. data fidelity) and the prior term. To verify the pliability of RGMs, we implement RGMs with diverse forward processes and regularization terms: 
\begin{itemize}[leftmargin=10pt, topsep=0pt]
\item 
We consider three prior knowledge by leveraging KLD, MMD, and DSWD, where each stands for the measurement of the difference between two distributions.
% KLD estimates the Jensen-Shannon divergence of two distributions
KLD measures how much two distributions diverge from each other entropically as introduced in Section \ref{sec:RGM}.
Using a kernel trick, MMD measures the mean squared difference of the statistics of two sets of samples. DSWD estimates the difference by calculating the sliced Wasserstein distance for two distributions for multiple projection vectors.
With the Gaussian noise forward process, we call these models \textit{RGM-KLD-D}, \textit{RGM-MMD-D}, and \textit{RGM-DSWD-D}, respectively. Here, the term ``D'' stands for ``denoising''. 
See Appendix B.2 for a detailed explanation.
\item Additional to the Gaussian noise forward process, which is primarily used in DDM, we also carry out an RGM-KLD with the multi-scale forward process introduced in \cref{sec:forward}.
In this case, the image is corrupted by a downsampling filter together with additive noise. Therefore, the model (termed by \textit{RGM-KLD-SR (naive)}) is demanded to conduct upsampling and denoising at the same time. 
To ease the task, we separate the downsampling and noising processes and perform them alternatively. We call a model using this separated schedule \textit{RGM-KLD-SR}. (See Appendix \ref{appen:shcedule} for details).
\vspace{-5pt}
\end{itemize}
\begin{figure}[t]
%   \vspace{-5pt}
  \begin{center}
    \includegraphics[width=0.45\textwidth]{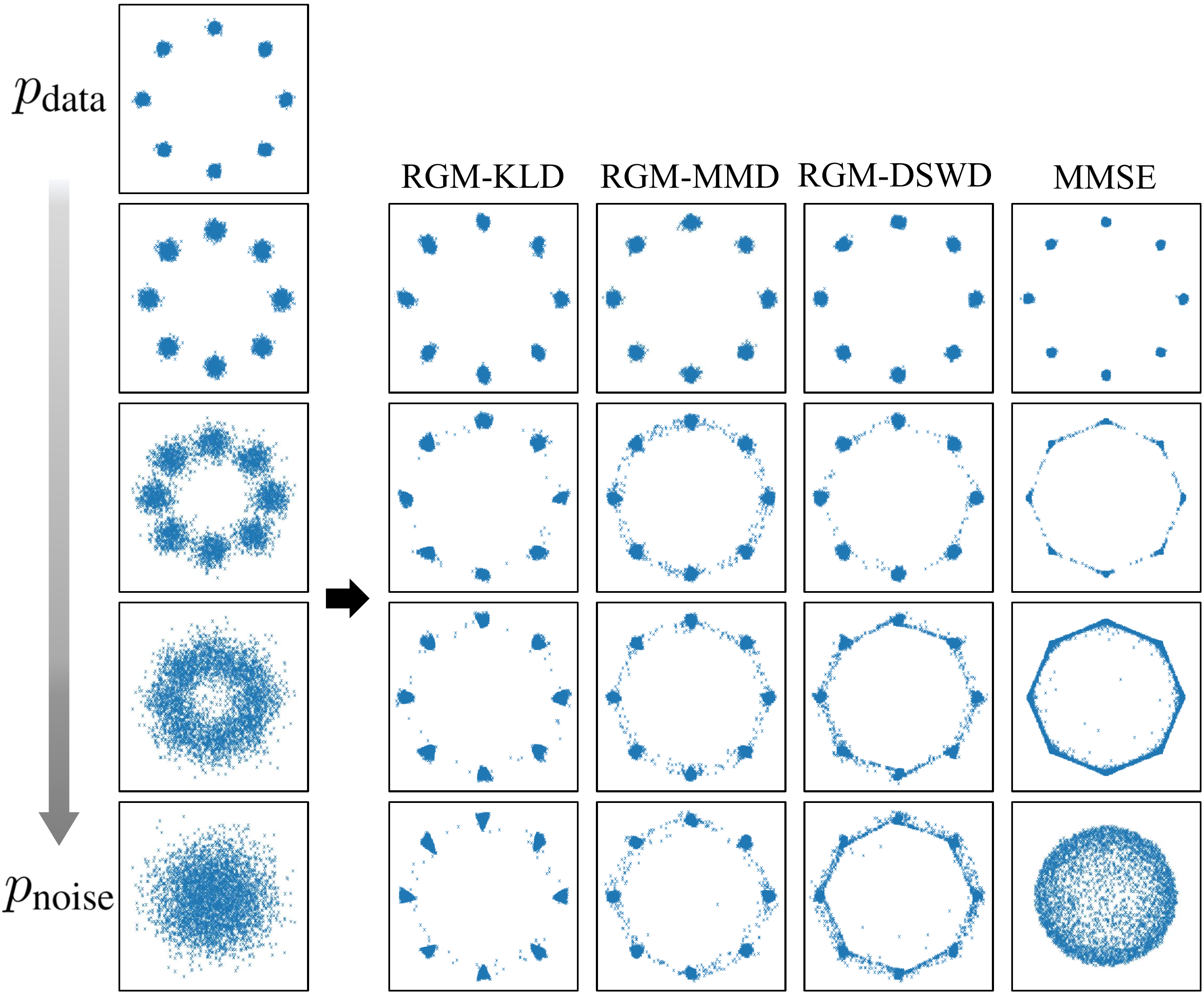}
  \end{center}
  \vspace{-5pt}
  \caption{Comparison of recovering density by MMSE versus three RGMs with different priors. All three RGMs are much more efficient than the MMSE approach.
%   Efficiency of MAP\myblue{-based approach} compared to MMSE. The prior term of our objective is variously parametrized with GAN, MMD, and DSWD. All three MAP\myblue{-based} approaches are much more efficient than MMSE approach, which shows the flexibility of how to parameterize the prior term of the RGMs.
  }
  \vspace{-15pt}
  \label{fig:gmm}
\end{figure}
\subsection{2D Toy Example} \label{sec:synthetic}
% We first employ
We first employ a two-dimensional example to validate the effectiveness and flexibility of our framework.
We adopt a mixture of Gaussian with eight components \cite{grathwohl2018ffjord} as a target distribution.
Our results are depicted in \cref{fig:gmm}.
As illustrated in the first column, we diffuse the data distribution through four different noise levels.
From left to right, each of the columns represents the learned distribution of our RGMs with 
 the regularization term parameterized by KLD, MMD, DSWD, and lastly MMSE estimation.
\vspace{-5pt}
\paragraph{Effectiveness of MAP-based approach}
\Cref{fig:gmm} shows the benefits of our methodology over the MMSE approach.
First, the rightmost column shows the failure of MMSE, where the modes of the distribution are connected and then missed.
This tendency exacerbates as the noise level increases.
Since the MMSE fails to reconstruct the data distribution even with a small rise in the noise level, the MMSE does not yield a satisfactory generative model with a small number of diffusion steps. Consequently, MMSE approaches, such as DDMs, require a large number of steps to stably recover the data distribution.
On the other hand, by adding the prior knowledge our RGMs generate samples from the multimodal distribution significantly better,
% We can see that the MAP-based approaches accurately estimate the density, 
which allow distribution recovery with a much smaller number of forward processes than the MMSE approach. This demonstrates the effectiveness of using the prior term $g$.
% Furthermore, we can observe the effect of the auxiliary variable $\bz$ by comparing the second and third columns in \cref{fig:gmm}. MAP with $\bz$ has higher sample quality, especially more pertinent to separate individual modes. The effect of $\bz$ is further amplified for more complex distributions, such as image data (See \cref{fig:z_lsun} and \cref{tab:ablation}).
% This synthetic experiment validates that our MAP approach enables accurate and efficient generative modeling.
\vspace{-17pt}
\begin{table}[t]
% \vspace{-5pt}
    \centering
    \caption{Results on unconditional generation of CIFAR10. } \label{tab:cifar10}
    % \vspace{2pt}
    \vspace{-5pt}
      \setlength\tabcolsep{1.0pt}
     \scalebox{0.73}{
    \begin{tabular}{ccccc}
    \toprule
    Class & Model &  FID ($\downarrow$) &   IS ($\uparrow$) &      NFE ($\downarrow$)         \\
    \midrule
    \multirow{3}{*}{\textbf{RGM}} &    RGM$-$DSWD$-$D & 3.11 & 9.08 & 4 \\ 
    & RGM-KLD-D\ \ \ &               3.04&    9.14 &     4            \\
&    RGM-KLD-SR&           2.47&       9.68 &    7            \\
    \midrule
    \multirow{19}{*}{\textbf{DDM}}  &  NCSN \cite{song2019generative}&      25.3&    8.87&     1000  \\
  &  DDPM \cite{ddpm}&      3.21&   9.46&     1000             \\
  &  Score SDE (VE) \cite{scoresde} &     2.20&      9.89&    2000             \\
  &  Score SDE (VP) \cite{scoresde}&       2.41&    9.68&    2000             \\
  &  Probability Flow (VP) \cite{scoresde}&  3.08&  9.83&    140             \\
  &  DDIM (50 steps) \cite{ddim}&           4.67&   8.78&     50             \\
  &  Recovery EBM \cite{recovery} &          9.58&  8.30&     180             \\
  &  LSGM \cite{vahdat2021score}&                2.10&     9.87&    147             \\
  &  FastDDPM ($T=50$) \cite{kong2021fast}&     3.41&     8.98&     50             \\
  &  VDM \cite{kingma2021variational}&                 4.00&     -   &    1000             \\ 
  &  UDM \cite{kim2021score}&                  2.33&   10.1&     2000             \\
  &  Gotta Go Fast \cite{jm2021ggf}&         2.44&   -   &    1000             \\
  &  Subspace Diffusion \cite{jing2022subspace} &      2.17& 9.94&    $\geq$ 1000     \\
  &  CLD \cite{dockhorn2021score} &                   2.25&  -   &     2000        \\
  &  DDGAN \cite{xiao2021tackling} &                3.75&    9.63&    4           \\
  & DEIS \cite{zhang2022fast} & 3.37 & 9.74 &  15\\
  & StyleGAN2+ES-DDPM \cite{lyu2022accelerating}  &  5.52 &  - & 101\\
  & DPM-Solver-3 \cite{lu2022dpm} & 2.70 & - & 30\\
  & GENIE \cite{dockhorn2022genie} & 3.94 & - & 20\\
    \midrule
  \multirow{5}{*}{\textbf{GAN}} & SNGAN$+$DGflow \cite{ansari2020refining} &           9.62&    9.35&  25             \\
    & AutoGAN \cite{gong2019autogan} &              12.4&    8.60&   1             \\
    &TransGAN \cite{jiang2021transgan} &            9.26&     9.02&    1             \\
    &StyleGAN2 w/o ADA \cite{karras2020training} &   8.32&     9.18&    1             \\
    &StyleGAN2 w/ ADA \cite{karras2020training} &       2.92&  9.83&    1             \\
    \midrule
    \multirow{4}{*}{\textbf{Others}} & NVAE \cite{vahdat2020nvae} &              23.5&        7.18&   1             \\
   & Glow \cite{kingma2018glow} &                48.9&   3.92&     1             \\
   & PixelCNN \cite{van2016pixel} &             65.9&   4.60&     1024             \\
   & VAEBM \cite{xiao2020vaebm} &             12.2&       8.43&    16             \\
    \bottomrule
  \end{tabular}}
%   \vspace{-5pt}
    \vspace{-10pt}
\end{table} 
\paragraph{Flexibility of the prior term}
% \paragraph{Validity of MAP approach}
Our RGMs have the freedom to parametrize the prior term $g$ of \eqref{eq:RGM_Dloss}. To demonstrate that the RGM framework universally works for variously parametrized prior terms, we manifoldly design the prior term by KLD, MMD, and DSWD.
 %in two additional ways other than GAN discussed in \cref{sec:RGM}: MMD and DSWD. 
 The results depicted in \cref{fig:gmm} validate that RGMs parametrized in three different ways show consistent performance, where they are all more efficient than the MMSE estimator. 
In particular, MMD measures the distance between
two distributions based on a pre-defined kernel, and hence $g$ is fixed rather than learned. Despite this simple structure, the results confirm
that our RGM with MMD is more efficient than MMSE.
% Comparing MMD with GAN and DSWD, 
% Furthermore, we can observe that RGMs in which the prior term and the generator are trained by matching each other (GAN, DSWD) deliver better performance than using the fixed prior term (MMD).

%We include experimental results on image distributions in \cref{sec:generation}, where a similar story continues.

\begin{figure}[t]
%   \vspace{-10pt}
 \centering
 \includegraphics[width=0.78\linewidth]{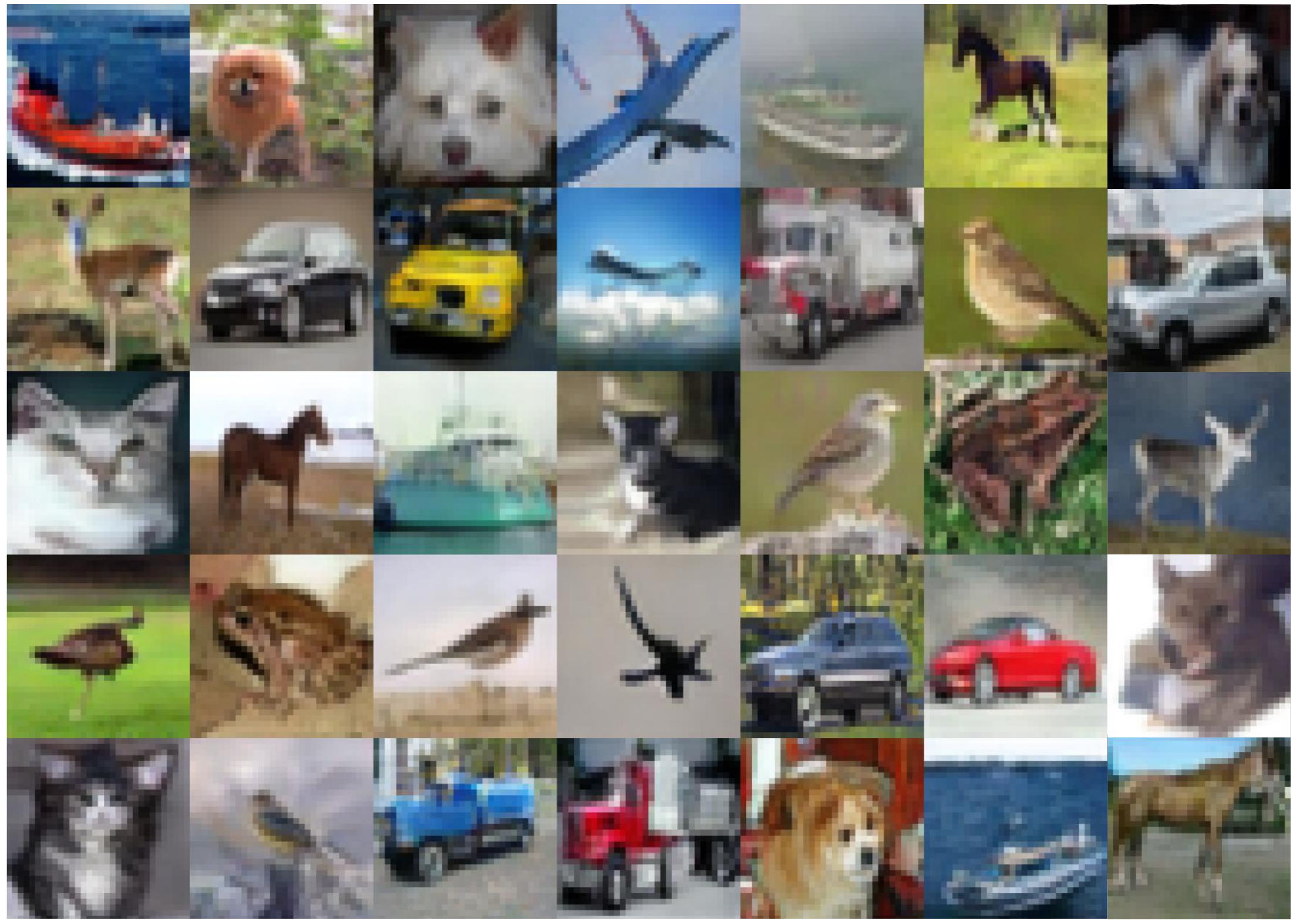}
 \vspace{-5pt}
\caption{CIFAR10 generated samples.}
\label{fig:cifar}
\vspace{-15pt}
\end{figure}
\subsection{Image Generation} \label{sec:generation}
We compare the performance of our RGMs with several existing baselines. 
% For quantitative comparison, 
We use Fr{\'e}chet Inception Distance (FID) and Inception Score (IS) as the evaluation metrics.
We also report
% measure the inference speed by 
the number of network function evaluations (NFEs). For DDMs and RGMs, NFE value and real inference time are proportional. Following \cite{scoresde, dockhorn2021score}, we focus on the widely used CIFAR10 unconditional image generation benchmark \cite{krizhevsky2009learning} and also validate the performance of RGMs on large-scale ($256\times 256$) images: CelebA-HQ \cite{liu2015deep} and LSUN Church \cite{yu2015lsun}.
\Cref{tab:cifar10,tab:celeba} summarize the quantitative evaluations on CIFAR10 and CelebA-HQ, respectively. The qualitative performance of RGM-KLD-D is depicted in \cref{fig:samples,fig:cifar}.
\vspace{-5pt}
\paragraph{Results} 
We can see that our models are comparable to the best existing DDMs on CIFAR10 and achieve the state-of-the-art FID score on CelebA-HQ-256.
% We can see that our model achieves the state-of-the-art FID score on CelebA-HQ-256. On CIFAR10, our models are comparable to the best existing DDMs and GAN models.
Although the best denoising models obtain better results than ours on CIFAR10, they use a much larger number of denoising steps (e.g. ScoreSDE with VESDE requires 2000 steps).
Notably, our RGM-KLD-SR achieves FID 2.47 and IS 9.68 with only seven steps, which is state-of-the-art sampling FID performance when NFE is limited.
% Notably, our RGM-KLD-SR$^{+}$ achieves FID 2.47 and IS 9.68 with only seven steps.
The overall results confirm that our method immediately eliminates the need for an expensive sampling scheme while still maintaining the density estimating capability of DDMs.
Interestingly, RGM-KLD-SR outperforms RGM-KLD-D by a large margin even with far fewer latent variables than RGM-KLD-D.
This improved performance may be attributed to the increase in NFE; however, the FID of RGM-KLD-D with $T=8$ reported in \cref{tab:ablation_appen} confirms that it is not.
In addition, RGM with the DSWD prior term retains comparable performance. This verifies that our MAP-inspired objective \eqref{eq:RGM_Dloss} works universally well, not being tied to how we parametrize the prior term.
% even when the prior term is parametrized in a way other than the GAN structure.
The overall results indicate that the prior knowledge regularized estimation of RGMs is a promising way of generating high-quality samples in limited steps.
More uncurated images can be founded in \cref{appen:samples}.
\begin{table}[h]
\vspace{-2pt}
    \centering
     \caption{Results on generation of CelebA-HQ-256. } \label{tab:celeba}
    \vspace{-2pt}
      \setlength\tabcolsep{7.0pt} %{10.0pt}
     \vspace{-5pt}
     % \scalebox{0.95}
    \scalebox{0.79}{
    \begin{tabular}{ccccc}
   \toprule
    Class & Model &                FID ($\downarrow$) &      NFE ($\downarrow$)         \\
    \midrule
    \textbf{RGM} & RGM-KLD-D&            \textbf{7.15} &    4            \\
    \midrule
    \multirow{5}{*}{\textbf{DDM}}
  &  Score SDE (VP) \cite{scoresde} &    7.23 &    4000             \\
  & Probability Flow \cite{scoresde} & 128.13 & 335 \\
  &  LSGM \cite{vahdat2021score}&       7.22 &    23             \\
  &  UDM \cite{kim2021score}&         7.16&     2000            \\
  &  DDGAN \cite{xiao2021tackling} & 7.64 &      4           \\
    \midrule
  \multirow{4}{*}{\textbf{GAN}} & PGGAN \cite{karras2017progressive} &   8.03 &    1            \\
    & Adv. LAE \cite{pidhorskyi2020adversarial} &  19.2  &   1             \\
    &VQ-GAN \cite{esser2021taming} &    10.2 &    1             \\
    & DC-AE \cite{parmar2021dual} &  15.8  &    1             \\
    \midrule
    \multirow{3}{*}{\textbf{VAE}} & NVAE \cite{vahdat2020nvae} &    29.7 &   1             \\
   & VAEBM \cite{xiao2020vaebm}&  20.4  &     1             \\
   & NCP-VAE \cite{aneja2021contrastive} &     24.8&      1         \\
    \bottomrule
  \end{tabular}}
    \vspace{-12pt}
\end{table}
\subsection{Ablation Studies} \label{sec:ablation}
This section is devoted to validating that the structure of the RGM framework is well-organized, with all parts of our objective, including fidelity term, prior knowledge, auxiliary variable, and regularization parameter, faithfully fulfilling their respective roles.
% This section is devoted to ablation analyses which show that all parts of our MAP objective, fidelity term, prior term, auxiliary variable, and regularization parameter, each play an important role in our performance of density estimation.
For a fair comparison, we used the same network for all experiments.
\vspace{-5pt}
\begin{figure}[t]
\centering
\begin{minipage}{.495\linewidth}
    \centering
 \begin{center}
    \includegraphics[width=0.99\textwidth]{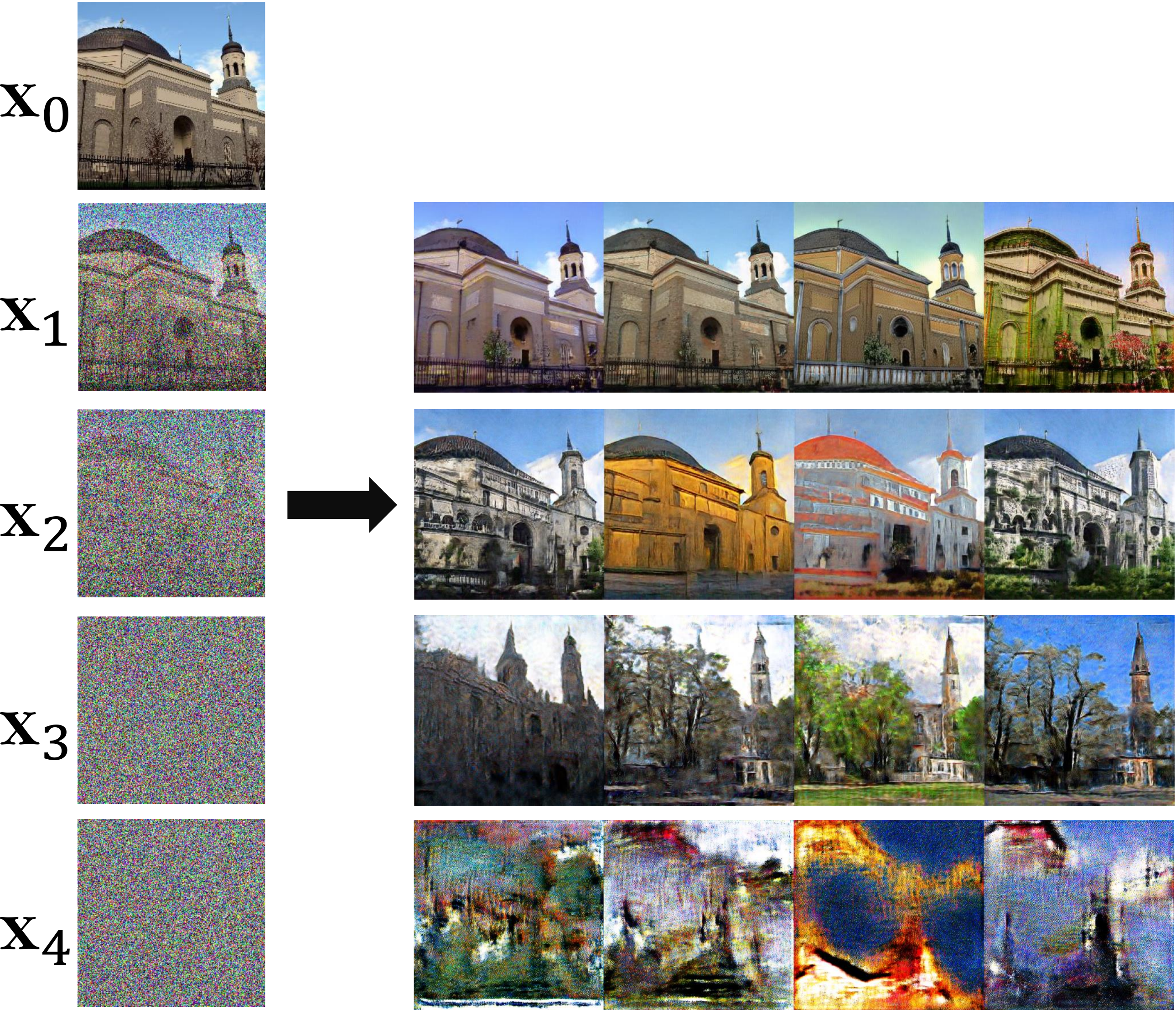}
  \end{center}
  \vspace{-5pt}
  \caption{Study on the effect of auxiliary variable $\bz$.} \label{fig:z_lsun}
\end{minipage}
\hfill
\begin{minipage}{.46\linewidth}
   \vspace{-0pt}
   \begin{center}
 \includegraphics[width=1.0\linewidth]{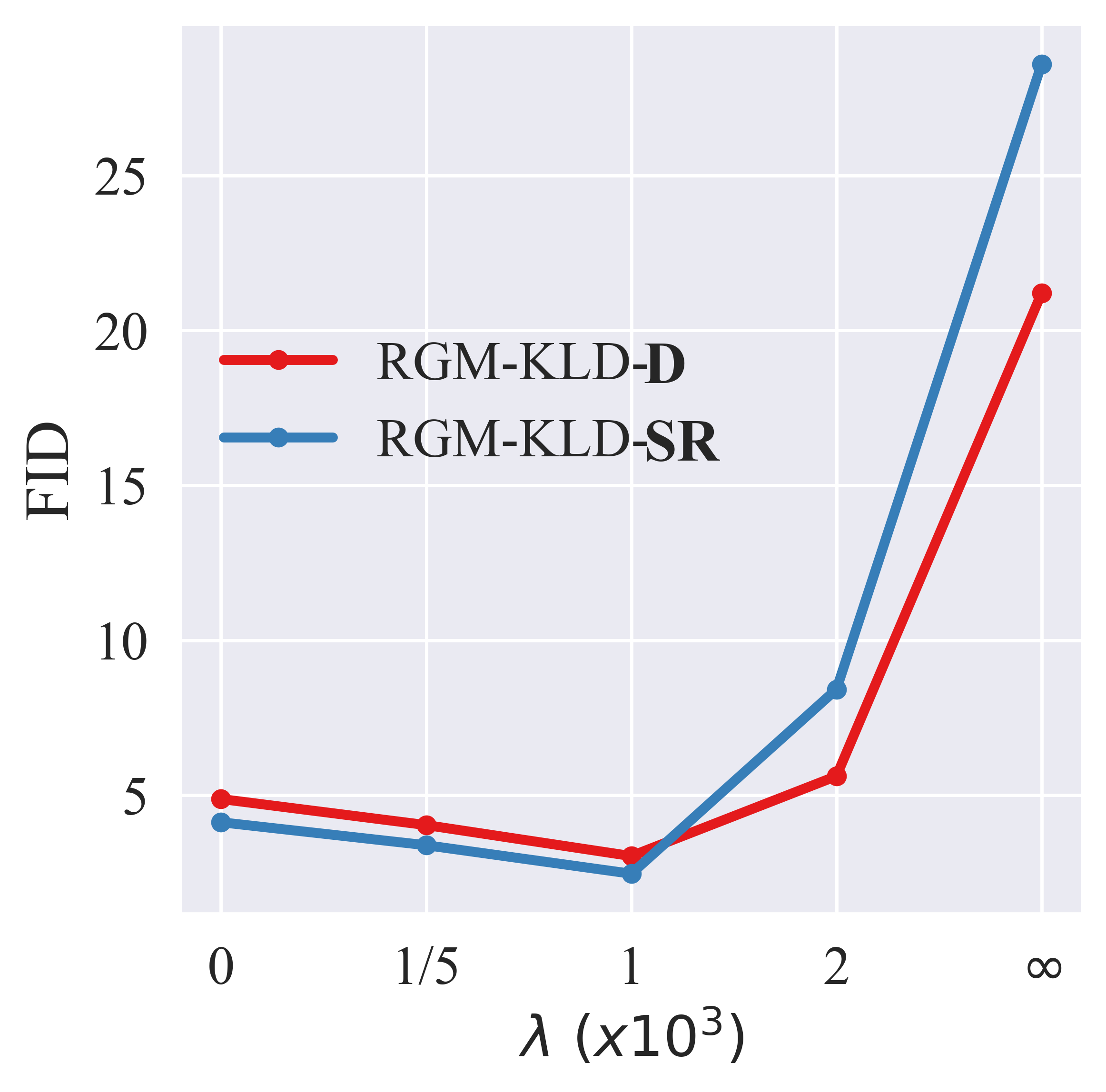}
 \end{center}
 \vspace{-13pt}
\caption{FIDs for different regularization parameter $\lambda$.}
\label{fig:lambda}
\end{minipage}
\vspace{-15pt}
\end{figure}
% \begin{figure}[h]
%   \vspace{-5pt}
%   \begin{center}
%     \includegraphics[width=0.49\textwidth]{figures/recover/z_lsun_main.pdf}
%   \end{center}
%   \vspace{-12pt}
%   \caption{Study on effect of $\bz$.} \label{fig:z_lsun}
%   \vspace{-25pt}
% \end{figure}
\begin{figure*}
%   \vspace{-15pt}
  \begin{center}
    \includegraphics[width=0.99\textwidth]{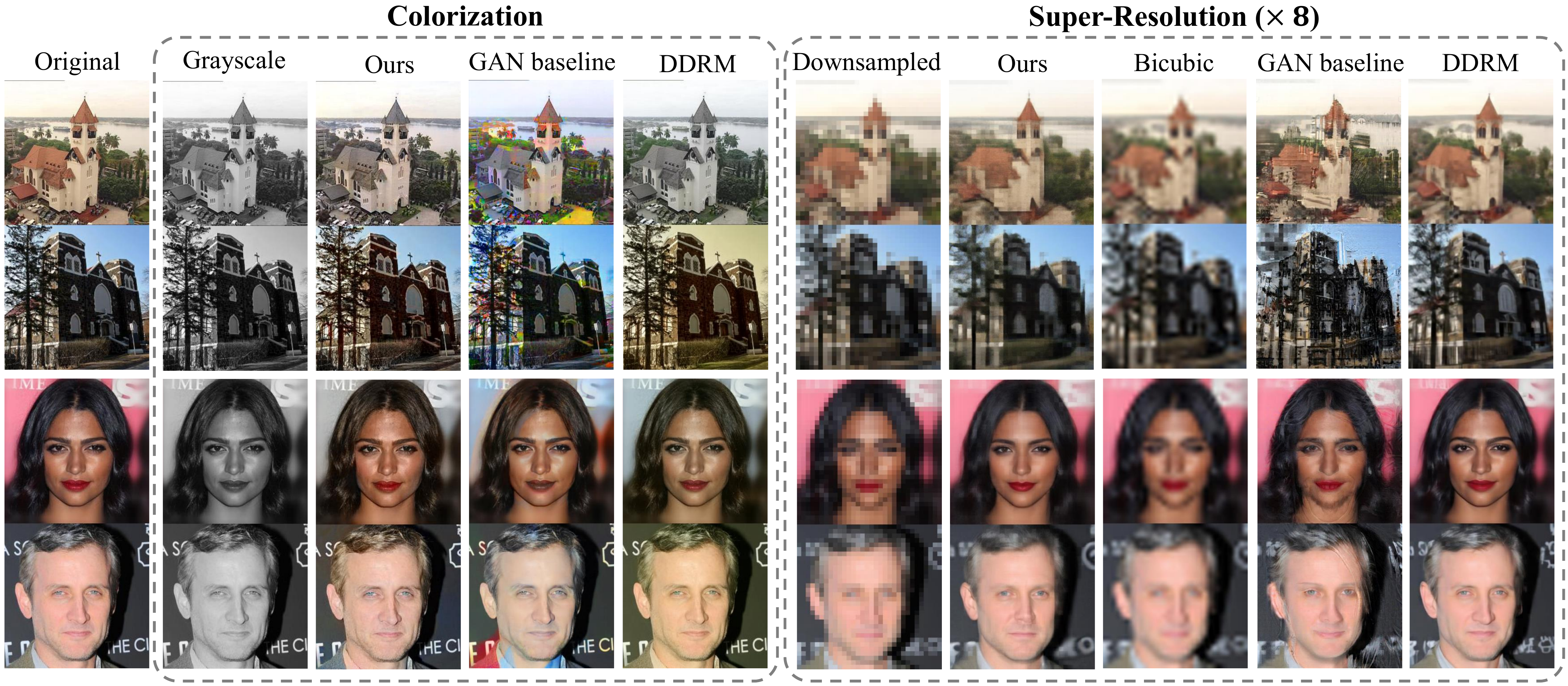}
  \end{center}
  \vspace{-10pt}
  \caption{Colorization (left) and super-resolution (right) results on LSUN and CelebA-HQ datasets.} \label{fig:color}
  \vspace{-10pt}
  \label{fig:inverse}
\end{figure*}
\paragraph{On the effect of Varying $\lambda$}
We investigate the sensitivity of the regularization parameter $\lambda$ in \eqref{eq:RGM_Dloss}.
Since it controls the relative importance between the fidelity term and the prior term, $\lambda$ is a trade-off hyperparameter that determines how much regularizes the joint distribution of $p_k$ and $p_{k+1}$.
In \cref{fig:lambda}, we present FID scores measured on CIFAR10 with the same number of degradation steps ($T=4$) and varying $\lambda$.
We can see that our models are quite robust with respect to $\lambda$.
An empirically observed sweet spot of $\lambda$ is $d/10\leq \lambda\leq d$ for the image size $d$, in which FID is no longer improved outside this threshold.
For small $\lambda$, the models put a lot of effort to recover the degradation, which hinders estimating data distribution.
Choosing a large $\lambda$ also results in performance degeneration.

\paragraph{On the Importance of Fidelity term}
The results of RGMs trained without the fidelity term also draw our attention. \Cref{tab:ablation} shows that the FID scores of both RGM-KLD-D and RGM-DSWD-D degenerate when there is no fidelity term. 
In particular, looking at RGM-DSWD-D, the performance without fidelity term is inferior to the vanilla DSWD model despite using multiple timesteps. 
This demonstrates that the performance improvement of our model is not solely due to the power of the existing generative models we used to design the prior knowledge.
On the other hand, including the fidelity term significantly improves performance.
In particular, RGM-DSWD-D achieves more than two times performance improvement over the vanilla DSWD.
The ablation studies we discuss here confirm that RGMs owe the performance improvement to the fidelity term, not simply because we borrow the expressivity from the regularization term.
% \textcolor{red}{The ablation studies we discuss here confirm that RGMs owe the performance improvement to the MAP-based estimation on which RGMs are built, not simply because we borrow the expressivity from the existing generative models.}
We further observe that with the help of the fidelity term, our model enhances the mode-collapsing resiliency of GAN (See \cref{fig:mode_collap}).
Overall results validate that the performance of RGMs owes to both fidelity and prior term, and the reliable regularization parameter should be determined to balance these two terms.

% There is another point that draws our attention. \textcolor{red}{Shown in Table \ref{tab:ablation}, in the case of using multiple timesteps, the performance is even worse than the original GAN or DSWD scheme if the fidelity loss is absent. Conversely, with the fidelity term equipped, FID score improves with superior performance for both RGM-KLD-D and RGM-DSWD-D.}
% % As shown in Table \ref{tab:ablation}, when there is no fidelity term on the objective, FID scores completely deteriorate. \textcolor{red}{The performance is even worse than the original GAN or DSWD scheme. However, }
% % In this case, the models are trained by the vanilla GAN loss. 
% % From the perspective of GAN, our MAP-based objective adds the fidelity term to the GAN loss function.
% We further observe that with the help of the fidelity term, our model enhances the mode-collapsing resiliency of GAN (See \cref{fig:mode_collap}).
% These results validate that the performance of RGMs owes to both fidelity and prior term, and the reliable regularization parameter should be determined to balance these two terms.
% \begin{figure}[h]
%   \vspace{-5pt}
%   \begin{center}
%     \includegraphics[width=0.42\textwidth]{figures/lambda.png}
%   \end{center}
%   \vspace{-15pt}
%   \caption{FIDs for different regularization parameter $\lambda$.} \label{fig:lambda}
%   \vspace{-20pt}
% \end{figure}
\begin{table}[h]
    \centering
    \caption{Ablation studies of RGMs on CIFAR10.} \label{tab:ablation}
    \vspace{-5pt}
      \setlength\tabcolsep{5.5pt} %{10.0pt}
     \scalebox{0.85}{
    \begin{tabular}{ccccc}
    \toprule
    Model \ \ \ \ \ \ & Multi-step & Fidelity & $\bz$ & FID ($\downarrow$)           \\
    \midrule
    &    \myxmark &    \myxmark  & \myxmark &     42.8 \\ %42.85   
    &    \myxmark &    \mycmark  & \mycmark &     14.6           \\  
\textbf{RGM-KLD-D}\ \ \ \ \ \ \ \ \ \      & \mycmark & \myxmark &  \mycmark &     32.5           \\

 & \mycmark & \mycmark & \myxmark & 3.87 \\
&    \mycmark &  \mycmark &  \mycmark &  3.04           \\
     \midrule
    
    & \myxmark &   \myxmark  & \myxmark &  7.12\\
\textbf{RGM-DSWD-D}\ \ \ \ \ \  &   \mycmark & \myxmark &    \mycmark &  16.3           \\
& \mycmark &  \mycmark &  \mycmark & 3.14           \\
\midrule
  \ \ \ \  \textbf{RGM-KLD-SR} (naive) &
    \mycmark &  \mycmark & \mycmark & 3.17           \\
    \ \ \ \  \textbf{RGM-KLD-SR}\ \ \ \ \ \ \ \ \ \ \ \ \  &    \mycmark &  \mycmark & \mycmark &  2.47           \\
    \bottomrule
  \end{tabular}
  }
    \vspace{-10pt}
\end{table} 
\vspace{-5pt}
\paragraph{On the role of $\bz$} 
%We sidestep the ill-posedness of \eqref{eq:inverse} by introducing random auxilliary variable $\bz$. To put forward some cogent reasons for introducing $\bz$, 
We include experimental results on LSUN church, which demonstrate how the auxiliary variable $\bz$ alleviates the ill-posedness of the inverse problem. By noising the upper-left image $\bx_0$, we obtain the forward trajectory $\{\bx_k\}_{k=1}^4$. The figures on the right are restored images of $\bx_k$ by RGM-KLD-D together with four different $\bz$.
We can see that reconstruction is almost unique when the noise level is small. But, as the noise level increases, a single $\bx_k$ has various reconstructions.
It is evident that assigning $\bz$ helps generate different denoised images from a heavily degraded $\bx_k$ through the guidance provided by $\bz$.
However, one might think that the ill-posedness is detoured by multi-step training using multiple $\sigma_k$ rather than through $\bz$. This claim can be refuted using the result of RGM-KLD-D without $\bz$ reported in \cref{tab:ablation}. We observe the significant difference in FIDs of RGM-KLD-D with and without $\bz$ under the same number of denoising steps, which indicates the effectiveness of $\bz$.
\vspace{-5pt}
\paragraph{On the forward process schedule}
Since the forward process determines the way of connecting the data and latent distributions,
it significantly affects the performance of models.
%\blue{the choice of the forward process schedule has significant practical implications.}
% it significantly affects the performance of models.
The first important factor is the number of forward steps $T$, which is directly related to NFE.
% one is forward step scheduling
In \cref{tab:ablation}, we ablate the effect of $T$.
% the number of forward steps.
When $T=1$, it may be difficult for the model to directly approximate the data distribution from the Gaussian noise. This is reflected in the poor FID score. 

We also study the forward process schedule of the SR model.
% The model to which this scheduling is applied is RGM-KLD-SR.
We can observe that the separation of the same forward process into two steps makes the model easier to learn, and this brings the performance enhancement of RGM-KLD-SR compared to RGM-KLD-SR (naive). From this, we would like to point out that properly designing the forward process can significantly increase performance. We leave the development of more useful and rigorous forward process as a promising future direction.
\begin{table*}[t]
    \centering
    % \vspace{-3pt}
    \caption{Quantitative comparison of RGM-KLD-D and RGM-KLD-SR on image reconstruction.} \label{tab:comparison_d_sr}
      \setlength\tabcolsep{9.2pt}
       \vspace{-5pt}
       \scalebox{0.95}{
    \begin{tabular}{ccccccccccccc}
    \toprule
    Model &        \multicolumn{4}{c}{Super-Resolution}       &  \multicolumn{6}{c}{Denoising} \\
     \cmidrule(lr){2-5}  \cmidrule(lr){6-11} 
        &   \multicolumn{2}{c}{($\times2$)} & \multicolumn{2}{c}{($\times4$)}& \multicolumn{2}{c}{($\sigma=10/255$)} & \multicolumn{2}{c}{($\sigma=20/255$)} & \multicolumn{2}{c}{($\sigma=40/255$)}  \\ 
        %  \cmidrule(lr){2-3}  \cmidrule(lr){4-5} \cmidrule(lr){6-7} \cmidrule(lr){8-9} \cmidrule(lr){10-11}
       &PSNR&SSIM&PSNR&SSIM&PSNR&SSIM&PSNR&SSIM&PSNR&SSIM \\
    \midrule
     RGM-KLD-D& 26.63 & 0.88 & 20.84 & 0.58 & \textbf{30.11} & \textbf{0.93} & \textbf{26.57} & \textbf{0.86} & \textbf{24.23} & \textbf{0.80} \\
    RGM-KLD-SR&\textbf{27.42} & \textbf{0.90} & \textbf{21.14} & \textbf{0.59} & 29.41 & 0.92 & 25.87 & 0.84 & 23.53 & 0.77 \\
    \bottomrule
  \end{tabular}}
    \vspace{-5pt}
\end{table*}
% The images from the Naive Sampler seem repetitive and they lack details.

\subsection{Inverse Problems} \label{sec:inverse}
While our model was originally devised to generate images, we further show the applicability of RGMs to inverse problems.
Recently, a promising approach in imaging inverse problems is to leverage a learned denoiser as an alternative to the proximal operator of splitting algorithms \cite{romano2017little, pnppgd}. Such methodology is referred to as Plug-and-Play (PnP) algorithms \cite{venkatakrishnan2013plug}. 
In a similar spirit, we utilize the trained RGMs as a modular part of the PnP algorithms to solve various inverse problems. 
In this section, we testify our RGM-KLD-D for two inverse problems; SR and colorization, by plugging our model into Douglas-Rachford Splitting algorithm \cite{lions1979splitting}. 
Details can be found in \cref{appen:pnp}.
\vspace{-5pt}
\paragraph{Results} We compare the performance of our model with current-leading models: We compare our model with DDRM \cite{kawar2022denoising}, which solves inverse problems with a pre-trained DDPM by a posterior sampling scheme. As a GAN baseline, we adopt StyleSwin \cite{zhang2022styleswin} and reconstruct the image by optimizing over the latent vector \cite{pan2021exploiting}.
We also consider bicubic interpolation as a baseline for super-resolution.
We observe that our model is capable of reconstructing faithful and realistic images, as evident in Figure \ref{fig:color}.
Compared with baselines, our model produces high-quality reconstructions across all the datasets. In particular, our model shows promising performance for colorization.
These results show the applicability of RGMs to PnP prior, and this will bring a range of potential applications, including image segmentation, conditional generation, and other imaging inverse problems. Additional quantitative and qualitative results are provided in \cref{appen:samples_inverse}.
\vspace{-5pt}
\paragraph{Comparison of RGM-D and RGM-SR}
We investigate the effect of the degradation process used during training on the performance of solving inverse problems.
We compare the reconstruction performance of RGM-KLD-D and RGM-KLD-SR which are trained on different degradation processes by applying both models to denoising and super-resolution (SR) tasks on CIFAR10. Quantitative results are presented in \cref{tab:comparison_d_sr}.
We can see that the RGM-KLD-SR that is trained based on SR actually performs the SR task better. Also, we can observe a similar tendency for denoising.
The results confirm that the degradation process used in training actually helps in solving the corresponding inverse problem.

\section{Related Work}
In recent years, DDMs \cite{ddpm, song2019generative, scoresde} have emerged as a class of density estimation models, first sparked by \cite{sohl2015deep}.
They define a sampling process as the reverse of a forward diffusion process that maps data to Gaussian noise by consecutively adding a small portion of the noise to the input data.
DDMs are known to faithfully estimate the data distribution and generate high-fidelity samples, however,
% As they faithfully estimate the data distribution and generate high-fidelity samples, they have rapidly been applied to various domains such as conditional generation \cite{lee2021priorgrad, ho2022cascaded}, audio synthesis\cite{kong2020diffwave, popov2021grad}, medical imaging \cite{song2021solving, chung2022score}, video generation \cite{ho2022video, yang2022diffusion}, and 3D point cloud generation \cite{lyu2021conditional}.
their major drawback is slow and expensive sampling speed. Many studies have been dedicated to circumventing this downside by developing a fast numerical solver \cite{gotta, zhang2022fast, tachibana2021taylor, liu2022pseudo} or using an alternative noising process such as non-Markovian \cite{ddim}, a second-order Langevin dynamics \cite{dockhorn2021score}, and non-linear diffusion processes \cite{de2021diffusion, chen2021likelihood}. 
Another line of work improves sampling efficiency by incorporating it into other generative models, including GAN \cite{xiao2021tackling, lyu2022accelerating}, and VAE \cite{vahdat2020nvae}.
\citet{xiao2021tackling} which enjoys small sampling steps by using GAN is one of our related works. 
%However, they did not introduce GAN from a MAP perspective and our model requires less training iteration to achieve the same performance. (See Appendix \ref{appendix:relation} for details.)
% Moreover, there have been distillation approaches \cite{salimans2022progressive, meng2022distillation}.
On a side note, all the aforementioned models use the Gaussian noising process as the forward process.

Recently, the literature has begun to replace the additive Gaussian noising process with other transforms. Breaking away from the diffusion process, \cite{rissanen2022generative} proposed a forward blurring process inspired by heat dissipation. They suggest a new generation process, but they specialize in the proposed blurring process and cannot be incompatible with other degradation processes.
Possibly the closest study to our work is Cold Diffusion \cite{bansal2022cold} which generalizes the diffusion process to arbitrary image transformations.
It seems to use a general transform similar to our models, but Cold Diffusion only uses deterministic degradation processes by entirely removing additive Gaussian noise, which hinders its density estimation performance.
Also, they use the MMSE objective, still requiring an array of several forward steps. We include a comparison with these related works in Appendix \ref{appen:comparison_related}.
% \vspace{-5pt}
\section{Conclusion and Future Work}
In this study, we presented a general framework for modeling efficient generative models through the lens of IR.
Compared to DDMs whose both forward and reverse processes are fixed to thousands of Gaussian steps, our approach provides more flexible models that eliminate expensive sampling and can enjoy versatile forward processes. 
We eliminated the usage of slow sampling by taking on the MAP-based approach and incorporating implicit priors. 
In addition, we propose a multi-scale method as an example of the usability of various forward processes. 
The experimental results showed that the image quality obtained was on par with the leading DDMs, and we achieved state-or-the-art performance using a limited number of forward steps.  We hope that this work provides a broad view of modeling useful generative models.

Our model has two degrees of freedom: One is how to parametrize the prior knowledge, and the other is the choice of the forward process. 
Designing new prior terms and degradation processes would be an interesting direction for future research.
%This opens up interesting directions for future research.
% In addition to the GAN we used, the prior term can be constructed in different ways.
% For example, the variational regularization can be explicitly parametrized via VAE.
%It could also be interesting to explore other degradation transformations.
Future work could include the comprehensive design of a convergence-guaranteed PnP algorithm for application to various inverse problems.
We leave these further extensions to future work.
Furthermore, notwithstanding the high performance, our methodology lacks theoretical justification.
We also leave this as interesting future work.

\section*{Acknowledgements}
This work was supported by the NRF grant [2012R1A2C3010887] and the
MSIT/IITP ([1711117093], [2021-0-00077], [No. 2021-0-01343, Artificial Intelligence Graduate School Program(SNU)]).
The author appreciate the financial support provided by the Data-driven Flow Modeling Research Laboratory funded by the Defense Acquisition Program Administration under Grant UD230015SD.
We are also grateful to Jaewoong Choi and Changyeon Yoon for reviewing an early draft of this paper and providing thoughtful feedback.

\nocite{langley00}

\bibliography{mybib}
\bibliographystyle{icml2023}

\newpage
\onecolumn
\appendix
\appendix

\section{Continuation of Related Works} \label{appendix:relation}
DDMs have been pertinent generative models by showing promising results on various generation tasks. 
DDMs degrade the data with a reference diffusion process and learn the data distribution by restoring it.
We have arranged DDMs in the context of restoration, and DDMs can be interpreted as an MMSE estimator for a denoising task.

\begin{itemize}[leftmargin=.12in, topsep=0pt]
\item Energy-based models (EBMs) are another line of generative models that learn the unnormalized data distribution by giving low energy to high-density regions in the data space.
As DDMs have demonstrated that recovery of a sequence of noisy data is more effective than directly approximating the data density, \citet{recovery} recently proposed a recovery energy-based model (REBM) by using a diffusion process. 
Inspired by DDMs, REBM learns a sequence of energy functions for the marginal distributions of the diffusion process.
More precisely, from the noisy observation $\tilde{\bx} =\bx+\xi,\ \xi\sim \mathcal{N}\left(0,\sigma^2 I\right)$, they estimate the conditional likelihood $p_\theta\left(\bx\mid\tilde{\bx}\right)\propto \exp^{-\mathcal{E}_\theta\left(\bx\mid\tilde{\bx}\right)}$ by learning the energy function
\begin{equation} \label{eq:recovery_energy}
\mathcal{E}_\theta=\frac{1}{2\sigma^2}\left\Vert\bx-\tilde{\bx}\right\Vert^2-f_\theta\left(\bx\right).
\end{equation}
They indeed learn the marginal density $f_\theta$ and infer the data through the recovery likelihood.
The marginal density $f_\theta$ is adversarially trained by assigning low energy to high-probability regions in the data space and high energy values outside these regions.
Since direct sampling from $p_\theta\left(\bx\mid\tilde{\bx}\right)$ is intractable, samples are usually drawn by leveraging Langevin dynamics (LD) \citep{neal1993probabilistic}, which is a conventional sampling method of EBMs.
Therefore, REBM trains marginal density $f_\theta$ using a kind of adversarial loss, but REBM is actually a MAP estimator implicitly defined by the sampling dynamics.
In other words, REBM learns the posterior distribution using the reference diffusion process, but it does not deviate from the traditional sampling method of EBM, still generating samples through inefficient LD.
There are two difficulties of such a  Markov
Chain Monte Carlo (MCMC) sampling: Applying MCMC in pixel space to sample one instance from the model is impractical due to the high
dimensionality and long inference time. As reported in \citep{xiao2021ebms}, the estimated density of EBMs can sometimes differ significantly from the data distribution, even if the model with the short-run LD produces relevant samples.
It is also known that the convergence of LD is very difficult when the energy function is complicated.

\item Another related work is a denoising diffusion GAN (DDGAN) \citep{xiao2021tackling}, which enjoys small sampling steps by using GAN.
DDGAN focuses on improving the sampling efficiency while maintaining the sample quality and mode coverage of DDMs.
The reason why DDMs adhere to the heavy sampling scheme is their common assumption that the true posterior is approximated by Gaussian distributions. This assumption holds only with small denoising steps.
When the number of denoising steps is reduced, the denoising distribution is no longer a Gaussian distribution, but a non-Gaussian multi-modal, which is usually intractable.
DDGAN breaks the Gaussian assumption by reducing the number of denoising steps and then approximates the non-Gaussian multimodal posterior distribution with the help of GAN.
DDGAN enhances the sampling efficiency of DDMs and also resolves the mode collapse problem of GANs by using a couple of denoising steps from the perspective of GAN literature.
The architecture of DDGAN is somewhat similar to that of our RGM-KLD-D.
However, there is a difference in a way of estimating MAP.
DDGAN assigns all responsibility for MAP estimation to the GAN structure.
On the other hand, our models learn the MAP-based estimator by separating the posterior distribution into the fidelity term and the prior term. Therefore, the model is much easier to learn than DDGAN.
As a consequence, RGM-KLD-D obtains substantial savings in terms of training iterations than DDGAN.
Specifically, in CIFAR10 experiments, DDGAN takes 400K iterations to achieve FID of 3.75. In comparison, our RGM-KLD-D only uses 150K iterations to achieve the same performance as DDGAN and takes 200K iterations for FID of 3.04.
For the CelebA dataset, DDGAN requires 750K iterations to attain FID 7.64, while RGM-KLD-D obtains the same FID score using only 450K iterations and FID 7.15 even with 500K iterations. Furthermore, our framework can be extended to various forward processes and regularization terms, which are more flexible and utilizable.
\end{itemize}
As such, there have been various density estimation models based on denoising.
Diffusion models, such as DDPM and score matching with Langevin dynamics and its variants, are MMSE-based estimators.
The model of REBM itself approximates the marginal density as we do, but our model is trained with MAP-based loss, whereas REBM generates samples from the posterior distribution through the sampling method.
Diffusion models and REBM train different estimators, but both models use a Langevin sampling scheme that requires thousands of network evaluations.
On the other hand, DDGAN is a model that can perform one-shot sampling with the help of GAN (away from the Langevin sampling), just like our RGMs.
However, since DDGAN learns the whole posterior density through the discriminator, it is more inefficient in terms of learning than our models, which separate the fidelity and the prior term.
Consequently, our RGMs achieve better performance than DDGAN with much fewer iterations.
All these models are restricted to the diffusion process.
Otherwise, our RGMs can enjoy flexible forward processes and are also given a degree of freedom in how to parametrize the prior term.
In other words, our approach does not need to restrict to the diffusion process and unlike DDGAN, which is limited to the GAN structure, it is possible to design the prior term by leveraging different generation models. This is further discussed in Appendix \ref{appen:train}.

\section{Implementation Details} \label{appen:implement_detail}
\subsection{Degradation Schedule} \label{appen:shcedule}
Let $\mathbf{A}_k$ and $\mathbf{\Sigma}_k$ be a degradation matrix and a noise variance on the $k$-th degradation step, respectively. 

Then, given a data $\bx$ sampled from the real data distribution $p_\text{data}$, a degraded data $\by_k$ on the $k$-th forward step is sampled from
\[p\left(\by_k\mid\bx\right)=\cN\left(\by_k;\mathbf{A}_k\bx,\mathbf{\Sigma}_k\right).
\]

\begin{figure}
    \centering
    % \vspace{-5pt}
    \includegraphics[page=1,width=0.90\textwidth]{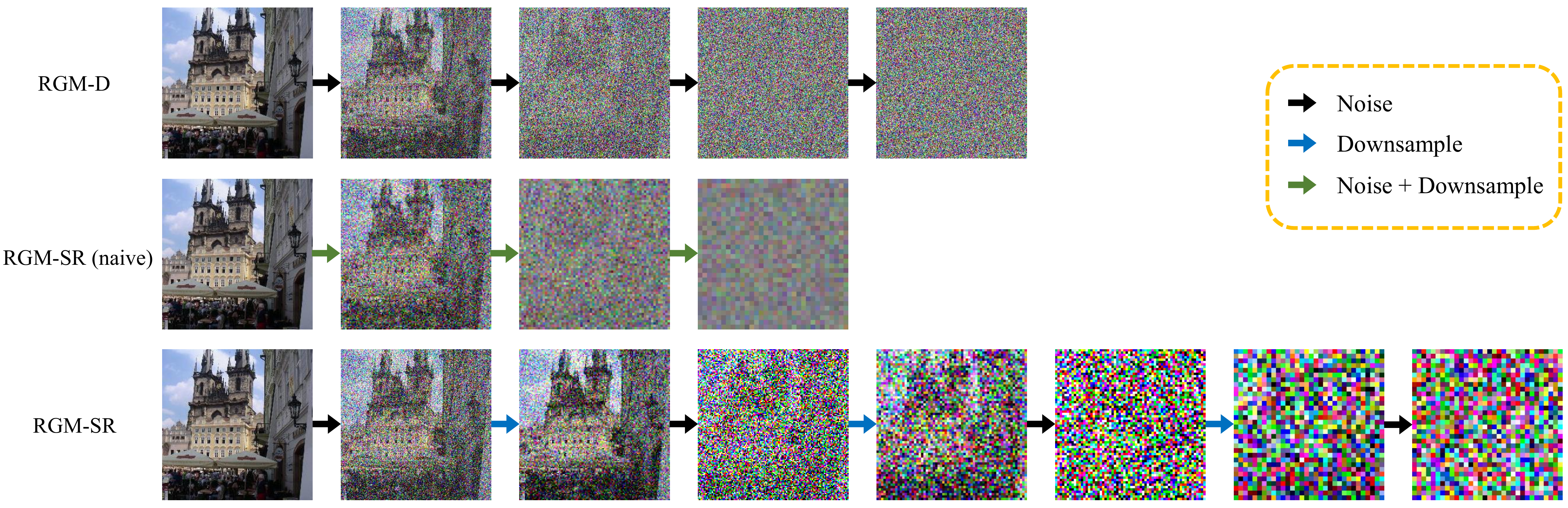}
    \caption{Degradation sequences for RGMs.}\label{fig:forward_process}
    % \vspace{-10pt}
\end{figure}

We denote the marginal distribution at the $T$-th degradation step as $p_T$.
Because our primary goal is to bridge $p_\text{data}$ to an easy-to-sample distribution $p_T$, (especially to a zero mean Gaussian distribution),
we gradually decrease the norm of $\mathbf{A}_k$ to zero as $k$ increases.
In Section \ref{sec:experiments}, we introduced two families of models based on the degradation schedule $\left\{(\mathbf{A}_k, \mathbf{\Sigma}_k)\right\}_{k=1}^T$ with the corner cases: RGM-KLD-D for $\mathbf{A}_k=\mathbf{I}$ and RGM-SR for $\mathbf{A}_k=\mathbf{P}_k$ a 2 × 2 averaging filter.
Roughly speaking, we consider three models based on different forward processes designed as follows:
\begin{itemize}[leftmargin=.32in, topsep=0pt]
\item RGM-D: $\text{noise} \rightarrow \text{noise} \rightarrow \text{noise} \rightarrow \text{noise} \rightarrow \cdots$.
\item RGM-SR (naive): $\text{downsample} + \text{noise} \rightarrow \text{downsample} + \text{noise} \rightarrow \cdots$.
\item RGM-SR: $\text{noise} \rightarrow \text{downsample} \rightarrow$  $\text{noise} \rightarrow \text{downsample} \rightarrow \cdots $,
\end{itemize}

shown schematically in \cref{fig:forward_process}.
With the following notations
\begin{align}
\beta_k &= \frac{1}{4} \left(\beta_{\rm{max}}-\beta_{\rm{min}}\right)\left(\frac{k}{T}\right)^2 + \frac{1}{2}\beta_{\rm{min}}\frac{k}{T}, \label{eq:beta}\\
\Tilde{\beta}_k &= \frac{1}{4} \left(\beta_{\rm{max}}-\beta_{\rm{min}}\right)\left(\frac{k}{T}\right)^4 + \frac{1}{2}\beta_{\rm{min}}\left(\frac{k}{T}\right)^2,
\end{align}
where $\beta_{\rm{max}}=20$ and $\beta_{\rm{min}} = 0.1$,
table \ref{tab:deg_schedule} details the explicit form of the forward processes used for each model.
\begin{table} [h]
    \centering
    \vspace{-0pt}
     \caption{The choice of schedule $\mathbf{A}_k$ and $\mathbf{\Sigma}_k$ and the corresponding latent distribution $p_T$ for RGM-D, RGM-SR (naive), and RGM-SR. $\mathbf{P}_k$ is a projection matrix that downscales the images by block averaging in a factor of $2^k$. For RGM-SR, we set $T$ in (\ref{eq:beta}) to be half of the total steps added by one.}
    \begin{tabular}{cccc}
    \toprule
       &       RGM-D                   &    RGM-SR (naive)         &   RGM-SR                          \\
    \midrule
    $\mathbf{A}_k$  &   $e^{-\beta_k} \mathbf{I}$          &  $e^{-\Tilde{\beta}_k} \mathbf{P}_k$ & 
    $e^{-\beta_{\lceil k/2\rceil}} \mathbf{P}_{\lfloor k/2 \rfloor}$ \\
   
    $\mathbf{\Sigma}_k$ & $\left(1-e^{-2\beta_k}\right)^2 \mathbf{I}$  &  $\left(1-e^{-2\Tilde{\beta}_k}\right)^2 \mathbf{P}_{k}^{\top} \mathbf{P}_{k}$ &  $\left(2^{\lceil k/2 \rceil}\left(1-e^{-2\beta_{\lceil k/2\rceil}}\right)\right)^2 \mathbf{P}_{\lfloor k/2 \rfloor}^{\top} \mathbf{P}_{\lfloor k/2 \rfloor}$  \\

    $p_T$ & $\cN\left(\mathbf{0}, \mathbf{I}\right)$ & $\cN\left(\mathbf{0}, \frac{1}{64}\mathbf{I}\right)$ & $\cN\left(\mathbf{0}, 4\mathbf{I}\right)$ \\
    \bottomrule
  \end{tabular}
    \vspace{-0pt}
    \label{tab:deg_schedule}
\end{table}
The noise schedule of RGM-D follows the Variance Preserving SDE provided by\citet{scoresde}, and others are implemented with a slight modification of them.

When we use the degradation matrix $\mathbf{A}_k$ as the averaging filter, the corresponding forward process downsamples the image while adding Gaussian noise. RGM according to this forward process, referred to as RGM-SR (naive), is demanded to super-resolve the degraded data while simultaneously denoising it.
It is considerably more difficult than the denoising task when the noise level is the same. 
To address this difficulty, we consider a newly scheduled degradation scheme that decomposes the forward process into downsampling and noising operations. We name the RGM designed in conjunction with this forward schedule as RGM-SR. 
As provided in Table \ref{tab:deg_schedule}, when the step $k$ is odd, the difference from the $(k+1)$-th step is only the projection matrix. Namely, only downsample is performed when sampling the $(k+1)$-th degraded data from the k-th degraded observation. Conversely, when $k$ is an even number, the forward process produces the $(k+1)$-th degraded image by adding the Gaussian noise. In summary, RGM-SR focuses on denoising the data in odd steps and super-resolving the data in even steps.
Provably due to the difficulty of performing super-resolution and denoising simultaneously, RGM-SR (naive) has the worst performance. Whereas RGM-SR, which uses the decomposed forward process, outperforms both RGM-D and RGM-SR by a large margin as reported in Section \ref{sec:generation}.

\subsection{Training RGMs} \label{appen:train}
In this section, we unambiguously elucidate how we train our RGMs. 
In Algorithms \ref{alg:train_post} and \ref{alg:train}, we summarize the two training procedures of GAN-based prior that are suited to different situations. 
We also provide an explanation of the training procedure of RGMs with other priors.
Moreover, the generation process is provided in Algorithm \ref{alg:Sample}.
%\begin{enumerate}[leftmargin=.32in, topsep=0pt]
%\item Train with posterior sampling
%\end{enumerate}

\paragraph{Training}
As proposed in Section \ref{sec:RGM}, RGMs learn the data distribution $p_\text{data}$ through the process of degrading the image through a forward process and then restoring it using the MAP-based objective \eqref{eq:RGM_Dloss}.
%RGMs learn the data distribution $p_\text{data}$ through the process of restoring the corrupted images $\by_k$, which are degraded by the forward process, using the MAP objective. 
However, since it is too difficult to restore the image directly from the Gaussian distribution in one shot, we use a handful of forward steps and train RGMs that recover the distribution between each step.
(We also include an ablation study on this in Appendix \ref{appen:ablation})
In other words, at each step $k$, we first sample a degraded image $\by_k$ of a given image $\bx\sim p_\text{data}$. The generator $G_\theta$ generates the restored image $\hat{\bx}$, and then, we degrade it by the posterior distribution $\hat{\by}_{k-1} \sim p\left(\hat{\by}_{k-1}\mid \by_k,\hat{\bx}\right)$.
We train our loss function so that $\hat{\by}_{k-1}$ becomes a restoration of $\by_k$.
The discriminator loss is also imposed on the $(k-1)$-th step.
Through the overall process, we ultimately learn the model that restores the distribution of the previous $(k-1)$-th step at each $k$-th step.
% By putting $\by_k$ and $\hat{\by}_{k-1}$ into our MAP loss function, we train our RGMs through the MAP objective that restores the distribution of the previous $(k-1)$-th step at each $k$-th step.

The training procedure is articulated in Algorithm \ref{alg:train_post}.

\begin{algorithm}
\caption{Training of RGMs with Posterior sampling}
\begin{algorithmic}[1]
\Require Dataset $\mathcal{D}$, degradation schedule $\{(\mathbf{A}_k,\mathbf{\Sigma}_k)\}_{k=0}^T$ with $\left(\mathbf{A}_0,\mathbf{\Sigma}_0\right)=\left(\mathbf{I},\mathbf{0}\right)$,
posterior distribution $p_{k\mid k-1}\left(\by_{k-1},\by_k\right)=\cN\left(\tilde{\mathbf{A}}_{k}\by_k,\tilde{\mathbf{\Sigma}}_{k}\right)$, generator $G_{\theta}$, discriminator $D_{\phi}$, and regularization parameter $\lambda \geq 0$.

\For{$i = 0, 1, 2, \dots$}
    \State Sample data $\bx \in \mathcal{D}$.
    \State Sample $k \sim {\rm Uniform}(\{1,2, \dots, T\})$.
    \State Sample $\bz \sim \mathcal{N}\left(0,\mathbf{I}\right)$.
    \State Sample degraded data $\by_k \sim \mathcal{N}\left(\mathbf{A}_k \bx, \mathbf{\Sigma}_k\right)$ and $\by_{k-1} \sim \mathcal{N}\left(\mathbf{A}_{k-1} \bx, \mathbf{\Sigma}_{k-1}\right)$.
    \State Generate an image $\hat{\bx} = G_\theta(\by_k, k, \bz)$.
    \State Degrade data by posterior sampling $\hat{\by}_{k-1} \sim p\left(\hat{\by}_{k-1}\mid \by_k,\hat{\bx}\right)$.
    \State Update $\phi$ by the following loss: $$\log \left(1-D_\phi \left(\hat{\by}_{k-1}, k-1\right)\right)+\log D_\phi \left(\by_{k-1}, {k-1}\right).$$
    \State Update $\theta$ by the following loss: $$\log\left(1 - D_\phi \left(\hat{\by}_{k-1}, k -1\right)\right) - \log D_\phi \left(\hat{\by}_{k-1}, k -1\right) +  \frac{1}{2\lambda} \left\lVert \left(\tilde{\mathbf{\Sigma}}^{\dagger}_{k} \right)^{\frac{1}{2}} \left(\tilde{\mathbf{A}}_{k}\hat{\by}_{k-1} - \by_{k}\right)  \right\rVert_2^2.$$
\EndFor
\end{algorithmic}
\label{alg:train_post}
\end{algorithm}

However, we can exactly formulate the posterior distribution only when the forward process satisfies certain conditions.
For all $k=1,\cdots,T$, if there exists $\left(\tilde{\mathbf{A}}_k, \tilde{\mathbf{\Sigma}}_k\right)$ satisfying
\begin{align}
    \mathbf{A}_k = \tilde{\mathbf{A}}_k \mathbf{A}_{k-1},\ 
    \tilde{\mathbf{\Sigma}}_k := \mathbf{\Sigma}_k - \tilde{\mathbf{A}}_k\mathbf{\Sigma}_k\mathbf{\tilde{A}^\top}_k \succ 0, \label{eq:Matrix}
\end{align}
we can explicitly construct a conditional distribution  $p_{k|{k-1}}(\by_k|\by_{k-1})=\mathcal{N}(\tilde{\mathbf{A}}_k\by_{k-1}, \tilde{\mathbf{\Sigma}}_k)$ and a posterior distribution \citep{ddpm,kingma2021variational,xiao2021tackling}.
For example, the forward process of RGM-D falls under this condition \eqref{eq:Matrix}, but that of RGM-SR does not. Therefore, the Algorithm \ref{alg:train_post} does not fit with RGM-SR.
To unravel such a problem, we propose a prevalent algorithm that is applicable to forward processes that are in discord with the condition \eqref{eq:Matrix}. See Algorithm \ref{alg:train}. 
The only difference from the Algorithm \ref{alg:train_post} is the replacement of the posterior sampling by the prior sampling in Line 5 and the data fidelity term in Line 9.
When the posterior distribution is unavailable, we corrupt the image $\hat{\bx}$ restored by the generator $G_\theta$ to the $(k-1)$-th degraded distribution using the $k$-th forward process rather than posterior sampling.
%we adjust the prior term $g$ to estimate the distribution of the previous degradation step, instead of directly recovering the data distribution.
%In other words, given the degraded data $\by_k$ of the data $\bx\sim p_\text{data}$, we first restore the data $\hat{\bx}=G_\theta\left(\by_k,k,\bz\right)\approx\bx$ and then we degrade the estimated image $\hat{\bx}$ to the $(k-1)$-th step $\by_{k-1}\sim\cN\left(\mathbf{A}_{k-1}\hat{\bx},\mathbf{\Sigma}_{k-1}\right)$.
Moreover, since the conditional distribution between $k$ and $(k-1)$ steps is unknown, we adopt the fidelity term of the image $\hat{\bx}$ reconstructed by the generator. This algorithm is universally applicable to general forward processes.
One notable fact is that RGM-D, whose posterior distribution is tractable, learns the data distribution better when using this algorithm than Algorithm \ref{alg:train_post}. This is discussed in detail in Appendix \ref{appen:ablation}.

\begin{algorithm}
\caption{Relaxed training algorithm of RGMs}
\begin{algorithmic}[1]
\Require Dataset $\mathcal{D}$, degradation schedule $\{(\mathbf{A}_k,\mathbf{\Sigma}_k)\}_{k=0}^T$ with $\left(\mathbf{A}_0,\mathbf{\Sigma}_0\right)=\left(\mathbf{I},\mathbf{0}\right)$, discriminator $D_{\phi}$, generator $G_{\theta}$, and regularization parameter $\lambda \geq 0$.
\For{$i = 0, 1, 2, \dots$}
    \State Sample $\bx \in \mathcal{D}$.
    \State Sample $k \sim {\rm Uniform}(\{1,2, \dots, T\})$.
    \State Sample $\bz \sim \mathcal{N}\left(0,\mathbf{I}\right)$.
    \State Sample degraded data $\by_k \sim \mathcal{N}\left(\mathbf{A}_k \bx, \mathbf{\Sigma}_k\right)$ and $\by_{k-1} \sim \mathcal{N}\left(\mathbf{A}_{k-1} \bx, \mathbf{\Sigma}_{k-1}\right)$.
    \State Generate an image $\hat{\bx} = G_\theta(\by_k, k, \bz)$.
    \State Degrade $\hat{\bx}$ by $\hat{\by}_{k-1} \sim \mathcal{N}\left(\mathbf{A}_{k-1} \hat{\bx}, \mathbf{\Sigma}_{k-1}\right)$.
    \State Update $\phi$ by the following loss: $$\log \left(1-D_\phi \left(\hat{\by}_{k-1}, k-1\right)\right)+\log D_\phi \left(\by_{k-1}, {k-1}\right).$$
    \State Update $\theta$ by the following loss: $$\log\left(1 - D_\phi \left(\hat{\by}_{k-1}, k -1\right)\right) - \log D_\phi \left(\hat{\by}_{k-1}, k -1\right) +  \frac{1}{2
    \lambda} \left\lVert \left(\mathbf{\Sigma}^{\dagger}_k \right)^{\frac{1}{2}} \left(A_{k}\hat{\bx} - \by_{k}\right)  \right\rVert_2^2.$$
\EndFor
\end{algorithmic}
\label{alg:train}
\end{algorithm}

\paragraph{Training with other priors}
Without being tied to the GAN discriminator, our RGMs have the freedom to parametrize the prior term $g$ of regularizer \eqref{eq:RGM_Dloss} in any other way. To demonstrate that the RGM framework universally works for variously parametrized prior terms, we design the prior term in two additional ways: \textit{maximum mean discrepancy (MMD)} \citep{dziugaite2015training} and \textit{distributed sliced Wasserstein distance (DSWD)} \citep{nguyen2020distributional}:
\begin{itemize}[leftmargin=.12in, topsep=0pt]
\item 
We replace KLD objective to MMD, a two-sample test based on kernel maximum mean discrepancy \citep{li2017mmd}.
For given two sets of data $X=\left\{x_1, x_2 \dots, x_M \right\}$ and $Y=\left\{y_1, y_2 \dots, y_M \right\}$, the MMD prior $g(X, Y)$, which estimates the MMD distance, is defined as follows;
\begin{equation}
    g\left(X,Y\right) = \frac{1}{{M\choose2}}\left[\sum_{i\neq j} k\left(x_i, x_j\right) - 2 \sum_{i\neq j} k\left(x_i, y_j\right) + \sum_{i\neq j} k\left(y_i, y_j\right)\right],
\end{equation}
where $k$ is a positive definite kernel. Following the prior works \citep{dziugaite2015training, li2015generative, li2017mmd}, we use a mixture of RBF kernels $k(x,x')=\sum_{i=1}^n k_{\sigma_i}(x,x')$ where $k_{\sigma}$ is a Gaussian kernel with bandwidth parameter of $\sigma$.

\item 
To measure the distance of two datasets $X=\left\{x_1, x_2 \dots, x_M \right\}$ and $Y=\left\{y_1, y_2 \dots, y_M \right\}$, sliced Wasserstein-based framework (SW) projects the data into a one-dimensional vector then explicitly calculates the Wasserstein distance on the projected space. In such an explicit calculation, SW can be freed from an unstable adversarial framework. Recently, \citet{nguyen2020distributional} has proposed a novel and efficient method to obtain useful projection samples, hence, we followed the implementation of this prior work in our experiments. Specifically, following \citet{nguyen2020distributional}, we use the learnable feature function and calculate DSWD on the feature space for CIFAR10 experiments. In other words, we replace the prior term $g$ of line 9 of Algorithm \ref{alg:train} to DSWD objective.
\end{itemize}

The results on the 2D synthetic example discussed in Figure \ref{fig:gmm} validate that RGMs parametrized in three different ways show consistent performance, where they are all more efficient than the MMSE estimator.
Furthermore, we also carried out the experiment of RGM-D with the DSWD prior, termed RGM-DSWD-D, on CIFAR10. Consequently, RGM-DSWD-D achieves an FID score of 3.14 retaining comparable performance with RGM-KLD-D. The overall results verify that our MAP approach works universally well for the various prior terms.

\begin{figure}
    \centering
    % \vspace{-5pt}
    \includegraphics[page=1,width=0.50\textwidth]{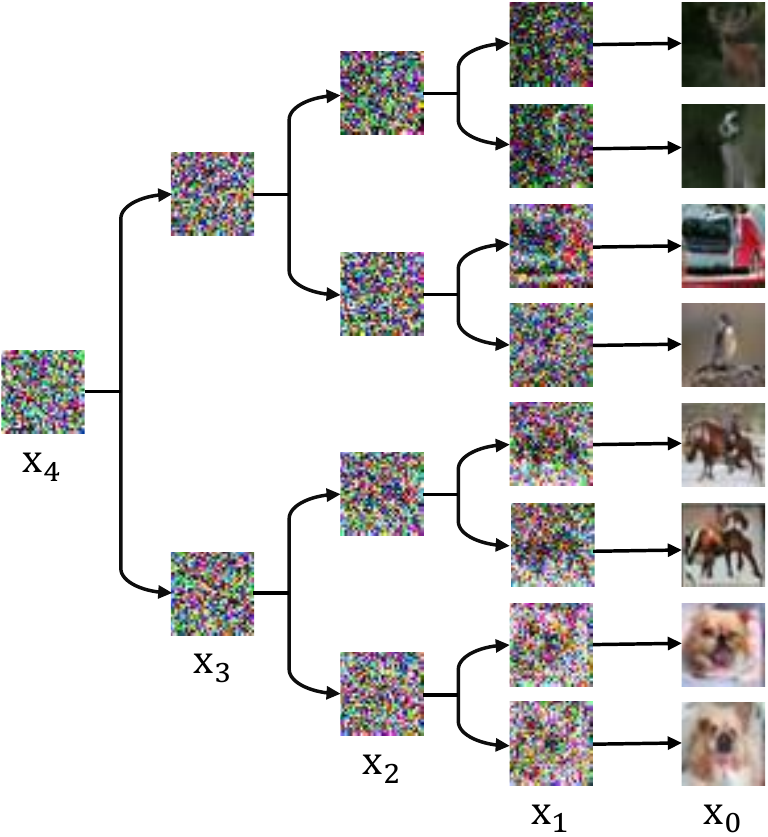}
    \caption{Hierarchical generation process of RGMs.}
    \label{fig:sampling}
    \vspace{-10pt}
\end{figure}

\paragraph{Sampling}
\begin{wrapfigure}{R}{0.6\textwidth}
\vspace{-10pt}
    \begin{minipage}{0.6\textwidth}
\begin{algorithm}[H]
\captionof{algorithm}{Sampling Procedure of RGMs}
\begin{algorithmic}[1]
\Require Trained generator $G_\theta$ and degradation schedule $\left\{\mathbf{A}_k, \mathbf{\Sigma}_k\right\}_{k=1}^T$.

\State Sample initial state $\by_T \sim \mathcal{N}\left(0, \mathbf{\Sigma}_T\right)$.
\For{$k = T-1, T-2, \dots, 0$}
    \State Sample $\bz \sim \mathcal{N}(\mathbf{0},\mathbf{I})$.
    \State Restore image $\hat{\bx}_{k}$ by $\hat{\bx}_{k} = G_\theta\left(\by_{k+1}, k+1, \bz\right)$.
    \State Sample $\by_{k}\sim \mathcal{N}\left(\mathbf{A}_k\hat{\bx}_{k}, \mathbf{\Sigma}_k\right)$.
\EndFor \\
\Return $\hat{\bx}_0$
\end{algorithmic}
\label{alg:Sample}
\end{algorithm}
\end{minipage}
\end{wrapfigure}
The sampling algorithm is summarized in Algorithm \ref{alg:Sample}.
Starting from a latent variable $\by_T\sim p_T$,
the trained $G_\theta$ generates the restored image $\tilde{\bx}=G_\theta\left(\by_{k+1},k,\bz\right)$ with a randomly selected auxiliary variable $\bz$ from the $(k+1)$-the degraded image $\by_{k+1}$, and then corrupt it by passing the $k$-th forward process. Continue this procedure until
$k=0$. When we train our model with Algorithm \ref{alg:train_post}, line 5 should be replaced by the posterior sampling.
For a schematic representation of this hierarchical sampling process of RGMs, see \cref{fig:sampling}.
% \begin{wrapfigure}{R}{0.4\textwidth}
% \centering
%     % \vspace{-5pt}
%     \includegraphics[page=1,width=0.40\textwidth]{figures/sampling.pdf}
%     \caption{Hierarchical generation process of RGMs.}
%     \label{fig:sampling}
%     \vspace{-10pt}
% \end{wrapfigure}
\subsection{Implementation Details}
We refer to \citep{nguyen2020distributional} for the precise definition of hyperparameters of RGM-DSWD-D.

\paragraph{Experiments on 2D dataset}
In the implementation of the two-dimensional Gaussian Mixture, we use a 3-layered MLP of 32 hidden dimensions for both generator and discriminator with Tanh activation. We concatenated all the inputs and passed them through the network.
For the RGM-KLD-D experiment, models are trained for 100K iterations with a learning rate of $10^{-4}$, a batch size of 1000.
For the RGM-DSWD-D experiment on the 2D data, we use the number of iterations of 100K, the number of projections of 10, 10 DSW iterations, and $\lambda_C=10$.
For the MMD experiment, we applied kernel bandwidths of 0.1, 0.5, 1, 2, and 10.
%For both 2D toy and CIFAR10 experiments, we use the same architecture and hyperparameters with RGM-D unless stated. 

\paragraph{Image generation}
To optimize our RGMs, we mostly followed the previous literature \citep{xiao2021tackling},
%For \green{image generation} experiments \green{on} CIFAR10, CelebA-HQ, and LSUN Church, we   follow experimental settings of DDGAN \citep{ddgan} 
including network architectures, $R_1$ regularization, and optimizer settings.
Note that our code is largely built on top of DDGAN \footnote{\url{https://github.com/NVlabs/denoising-diffusion-gan}} (MIT License).
We vary the discriminator by simply changing input channels into three.
Moreover, we use a learning rate of $2\times 10^{-4}$ for generator update in all experiments and a learning rate of $10^{-4}$ for discriminator update.
We use $\lambda^{-1}= 10^{-3}$ for image size of $32$, and $\lambda^{-1} = 5\times10^{-5}$ for image size of $256$.
The models are trained with Adam \citep{kingma2014adam} in all experiments. 
In CIFAR10 experiments, we train RGM-KLD-D and RGM-KLD-SR (naive) for 200K iterations and RGM-KLD-SR for 230K iterations.
Moreover, for RGM-DSWD-D implementation on CIFAR10, we use the output of the fifth convolutional layer of the discriminator as a feature vector. 
We use the number of iterations of 150K, the number of projections of 1000, 10 DSW iterations, and $\lambda_C=1$ for the DSWD experiment.
Lastly, we train RGM-KLD-D for 500K iterations and 300K iterations in LSUN experiments. 

\paragraph{Other details}
We train our models on CIFAR-10 using 4 V100 GPUs. The training takes approximately 40 hours on CIFAR-10. Moreover, the sampling of 100 samples takes approximately 0.25 seconds for RGM-KLD-D on single V100 GPUs.
For evaluation on CIFAR10, we use 50K generated samples to measure IS and FID. For CelebA-HQ-256, we use 30K samples to compute FID.

\subsection{Solving Inverse Problems} \label{appen:pnp}
Modern image processing algorithms reconstruct the ground-truth image by solving the following minimization problem:
\[
\underset{\bx}{\text{minimize}}\ {f_{\by}\left(\bx\right)+\lambda g\left(\bx\right)},
\]
where $f$ measures the fidelity to a corrupted observation $\by$, and $g$ constrains the solution space by measuring the complexity or noisiness of the image. 
Many imaging inverse problems, such as colorization, super-resolution (SR), and deblurring, fall under this form.
Since the above optimization problem does not have a closed-form solution in general, first-order proximal splitting algorithms, including half-quadratic splitting (HQS) \citep{geman1995nonlinear}, alternating direction method of multipliers (ADMM) \citep{boyd2011distributed}, solve the problem by operating individually on $f$ and $g$ via the proximal operator \citep{parikh2014proximal}.
With the aid of the emergence of deep learning, Plug-and-Play (PnP) algorithms \citep{venkatakrishnan2013plug} have recently begun to 
connect proximal splitting algorithms and deep neural networks by replacing the proximity operator of the regularization term $g$ with a generic denoiser \citep{romano2017little, reehorst2018regularization}.

Similarly, our trained RGMs can be used as PnP priors. In Section \ref{sec:inverse} we solved two inverse problems, colorization, and super-resolution, by plugging the trained RGMs into Douglas-Rachford Splitting (DRS) algorithm \citep{lions1979splitting}, following \citep{pnpadmm}. This is summarized in Algorithm \ref{alg:inverse}.
Starting from the degraded observation $\by$, the DRS algorithm updates the solution by alternatively utilizing proximal operations for both $f$ and $g$.
By iteratively updating the solution, the solution lies far outside the distribution on which our denoiser $G_\theta$ trained. For this out-distribution data, $G_\theta$ cannot recover the original image distribution, which in turn prevents the DRS algorithm from convergence.
To remedy this problem, the input of $G_\theta$ should always be within the trained distribution. Therefore, we push the updated solution into the learned distribution through the forward process.
Note that the proximal operation is calculated by utilizing efficient singular value decomposition proposed in \citet{kawar2022denoising}.
\begin{algorithm}[H]
\caption{Solving Inverse Problems by RGMs} \label{alg:inverse}
\begin{algorithmic}[1]
\Require A degraded observation $\by$, fidelity loss function $f_\by$, repeat number $M$, update rate $\alpha \in (0, 1]$, regularization parameter $\lambda \geq 0$, trained generator $G_{\theta}$, and degradation schedule $\left\{\mathbf{A}_k, \mathbf{\Sigma}_k\right\}_{i=1}^T$.

\State Initialize $\bx_K = \by$.
\For{$0, 1, \dots, M$}
    \For{$i = K, K-1, \dots, 1$}
        \State Sample $\hat{\by} \sim \mathcal{N}\left(\mathbf{A}\bx_i, \mathbf{\Sigma}_i\right)$ and $\bz \sim \mathcal{N}\left(0, \mathbf{I}\right)$.
        \State $\hat{\bx} \leftarrow G_\theta (\hat{\by}, i-1, \bz)$.
        \State $\hat{\bx} \leftarrow (1-\alpha)\bx_i+\alpha \hat{\bx}$.
        \State  $\Delta{\bx}\leftarrow {\rm prox}_{\lambda f_{\by}}(2\hat{\bx}-\bx_i) - \hat{\bx}$.
        \State $\bx_{i-1} \leftarrow \bx_i + \Delta{\bx}$.
    \EndFor
\EndFor \\
\Return $\bx_0$
\end{algorithmic}
\label{alg:Restoration}
\end{algorithm}

\paragraph{Settings \& Hyperparameters}
In SR experiments, we downscale images by using a block averaging filter by $r$ in each axis.
The filter is applied for the stride of $r$.
We experiment on $r=4$ and $r=8$ for LSUN and CelebA-HQ datasets.
In the CIFAR10 experiment, we use $r=2$ and $r=4$.
In colorization experiments, we simply degrade color images to gray by averaging images along the channels of each pixel.
All tasks are evaluated on hundred samples that are sampled from the evaluation dataset.
Table \ref{tab:hyper_inverse} reports the exact set of hyperparameters that we used in our experiments.
We set $K=2$ for colorization and $K=1$ for denoising and SR tasks.

On CIFAR10 experiments, to fairly compare RGM-KLD-D and the naive version of RGM-KLD-SR, we train both models with the same degradation steps of three ($T=3$). For RGM-KLD-D, we used $\mathbf{A}_k = e^{-\tilde{\beta}_k}\mathbf{I}$ and $\mathbf{\Sigma}_k = \left(1-e^{-2\tilde{\beta}_k}\right)^2 \mathbf{I}$. For RGM-KLD-SR, we used $\mathbf{A}_k = e^{-\tilde{\beta}_k}\mathbf{P}_k$ and $\mathbf{\Sigma}_k = \left(1-e^{-2\tilde{\beta}_k}\right)^2 \mathbf{P}_k^\top \mathbf{P}_k$.

\begin{table} [h]
    \centering
    % \vspace{-5pt}
      \setlength\tabcolsep{5.0pt}
       \caption{Hyperparameters used for solving inverse problems.} \label{tab:hyper_inverse}
       \vspace{-5pt}
       \scalebox{0.85}{
    \begin{tabular}{ccccccccc}
    \toprule
    %  \cmidrule(lr){2-5}  \cmidrule(lr){6-11} 
        &   \multicolumn{5}{c}{CIFAR10} & \multicolumn{3}{c}{LSUN/CelebA-HQ} \\ 
         \cmidrule(lr){2-6}  \cmidrule(lr){7-9}   
       &SR$\left(\times 2\right)$&SR$\left(\times 4\right)$&$\sigma=10/255$&$\sigma=20/255$&$\sigma=40/255$&SR$\left(\times 4\right)$&SR$\left(\times 8\right)$&Color \\
    \midrule
    M&          5&10&10&20&10&          40&40&20\\
    $\lambda$&  0.2&0.1&0.01&5&5&       10&10&5\\
    $\alpha$&   0.2&0.2&0.2&0.1&0.1&    0.05&0.05&0.5   \\
    \bottomrule
  \end{tabular}}
    \vspace{-10pt}
\end{table}
% We set NFE to 20 and $\alpha=0.1$.
% Moreover, in CIFAR10 $4\rm x$ super-resolution experiments, we set $K=2$.
% In LSUN and CELEBA-HQ experiments, ??.

\paragraph{Baselines}
We employed two main comparison models, namely DDRM \citep{kawar2022denoising} and GAN baseline, which is close to our work. Similar to our method, both comparison models assume that a degradation matrix is given and they iteratively update degraded images by using their knowledge obtained from the pretrained network and degradation matrix.
Moreover, our model and these comparisons do not require heavy additional training.
The implementation of DDRM follows its original implementation.
The implementation of GAN baseline mainly follows the implementation of DGP \citep{pan2021exploiting}, however, instead of using BigGAN \citep{biggan}, we replaced it with a pretrained model of StyleSwin \citep{zhang2022styleswin}, which is one of the state-of-the-art. For discriminator loss of DGP, we used the last feature vector of StyleSwin discriminator. We additionally adjusted the weights of the losses. For experiments in SR, we use an MSE loss weight of 1.0 and a discriminator loss weight of 1.0. For colorization, we use MSE loss weight of 1.0 and discriminator loss of 1.0 for the previous 400 iterations and 0.1 after that.
Other hyperparameters of GAN baseline implementation follow \citet{pan2021exploiting}.
We also compare our model with SDEdit \cite{meng2021sdedit}, a stroke-based diffusion model. In the implementation of SDEdit, we use total denoising steps of 200 with the number of repeats of three.

\section{Additional Results}

\subsection{Additional Ablation Studies} \label{appen:ablation}
In this section, we include additional ablation studies on our training procedure and the forward process schedule. 
All experiments are conducted on the CIFAR10 dataset and focused on RGM-KLD-D.
\paragraph{Directly restoring the data distribution}
\begin{wraptable}{r}{0.35\textwidth}
    \centering
    \vspace{-10pt}
     \caption{Additional ablation studies on CIFAR10 experiments.} \label{tab:ablation_appen}
     \vspace{-5pt}
    \scalebox{0.9}{
    \begin{tabular}{cccc}
    \toprule
    Model &                             FID ($\downarrow$)          \\
    \midrule
    Directly matching data&   21.2     \\
    RGM-KLD-D w/ posterior &     3.52           \\
    RGM-KLD-D\ \  $\left(T=8\right)$ &          6.50            \\
    \midrule
    RGM-KLD-D $\left(T=4\right)$ &    3.04           \\
    \bottomrule
  \end{tabular}
  }
    \vspace{-5pt}
\end{wraptable} 

%As we proposed in Section \ref{sec:method}, RGMs learn the data distribution $p_\text{data}$ through the process of restoring the corrupted images $\by_k$, which are degraded by the forward process, using the MAP objective.
Given a $k$-th degraded image $\by_k$, the generator is trained to restore the original image in one shot. Therefore, we can train RGMs to directly restore the real image distribution from each degraded step $k$.
%Given a $k$-th degraded image $\by_k$, the generator restores the image $\hat{\bx}\coloneqq G_\theta\left(\by_k,k,\bz\right)\sim p_g$, and the generator distribution $p_g$ is trained by MAP objective \eqref{eq:RGMloss} so that it approximates the real data distribution $p_g\approx p_\text{data}$. 
RGM-KLD-D trained in this say is denoted by \textit{Directly matching data} in Table \ref{tab:ablation_appen}. This model was trained in the same forward process as RGM-KLD-D ($T=4$).
% In the Table \ref{tab:ablation_appen}, \textit{Directly matching data} denotes the RGM-D trained with the MAP loss functional which restores the true data distribution from the corrupted images for each $k$. 
The FID score shows that the model has difficulties in learning the data distribution, falling short of FID score by 21.2. It seems that it is still difficult to directly restore the image of the real data distribution from a severely degraded image $\by_k$ ($k\approx T$) even with the help of auxiliary variable $\bz$.

\paragraph{Training with posterior sampling}
% On the other hand, because we learn the data density through a sequence of denoising processes, it is natural to extend our loss to estimate the distribution of the previous degradation step $k-1$.
While training, there are two ways to sample $\hat{y}_{k-1}$ from $\hat{y}_{k}$ (See line 7 of Algorithm \ref{alg:train_post} and  \ref{alg:train}). 
% As introduced in \ref{appen:train}, there are two ways to push the image, reconstructed by the generator, to the $(k-1)$-th degraded distribution; prior sampling and posterior sampling.
The posterior sampling (line 7 of Algorithm \ref{alg:train_post}) is theoretically well-grounded since it minimizes the statistical MAP loss of the posterior distribution.
However, to obtain an explicit form of posterior sampling, the forward process should be constrained to satisfy the conditions (\ref{eq:Matrix}).
Since the noising forward process of RGM-KLD-D satisfies these conditions, we trained RGM-KLD-D with both posterior sampling (Algorithm \ref{alg:train_post}) and prior sampling (\Cref{alg:train}) under the same setup.
In \cref{tab:ablation_appen}, \textit{RGM-KLD-D ($T=4$)} and \textit{RGM-KLD-D w/ posterior} refer to the model trained with prior and posterior sampling, respectively. 
As shown in \cref{tab:ablation_appen}, both models achieve similar results in terms of FID score, where RGM-KLD-D with prior sampling slightly precedes posterior sampling.
This verifies that the two training objectives of \cref{alg:train_post,alg:train} are somewhat consistent.
%As shown in Table \ref{tab:ablation_appen}, RGM-D with prior sampling slightly precedes posterior sampling in FID score. 
Because the performance is a bit better, we adopt the prior sampling in all our experimental studies.

\paragraph{Effect of the number of forward steps}
The number of forward steps is one of the important factors affecting the performance of the model. We investigated this in Section \ref{sec:ablation} by comparing a four-step model RGM-KLD-D ($T=4$) with the RGM-KLD-D ($T=1$), where we use only one degradation step. As reported in Table \ref{tab:ablation}, RGM-KLD-D ($T=1$) struggles to learn the data distribution because it needs to recover the real data distribution directly from Gaussian noise with one chance. On the other hand, RGM-KLD-D ($T=4$) estimates the data density well. Besides, what happens when we use more steps? Since our RGMs learn the data distribution in a way that restores the distribution of
the previous degradation step $(k-1)$ distribution from the $k$-th degraded distribution, one may expect that the models will be easier to estimate the density as the distribution between the two steps is closer by dividing the forward process with more steps. However, the opposite results are presented in Table \ref{tab:ablation_appen}. The results show that RGM-KLD-D ($T=8$) attains a higher FID score. In other words, dividing the forward process into smaller pieces does not enhance the model performance. In addition, this phenomenon is also observed for Directly matching data. RGM-KLD-D ($T=1$) can actually be regarded as Directly matching data ($T=1$), whereas the Directly matching data presented in the table uses $T=4$. Comparing these two, we can observe that the model using fewer degradation steps performs better.
% We conjecture that this tendency may come from
\citet{xiao2021tackling} reported a similar tendency. 
Choosing appropriate $T$ is crucial for algorithmic performance, but not straightforward how many steps are optimal.

\begin{figure}
  % \vspace{-5pt}
  \begin{center}
    \includegraphics[width=0.9\textwidth]{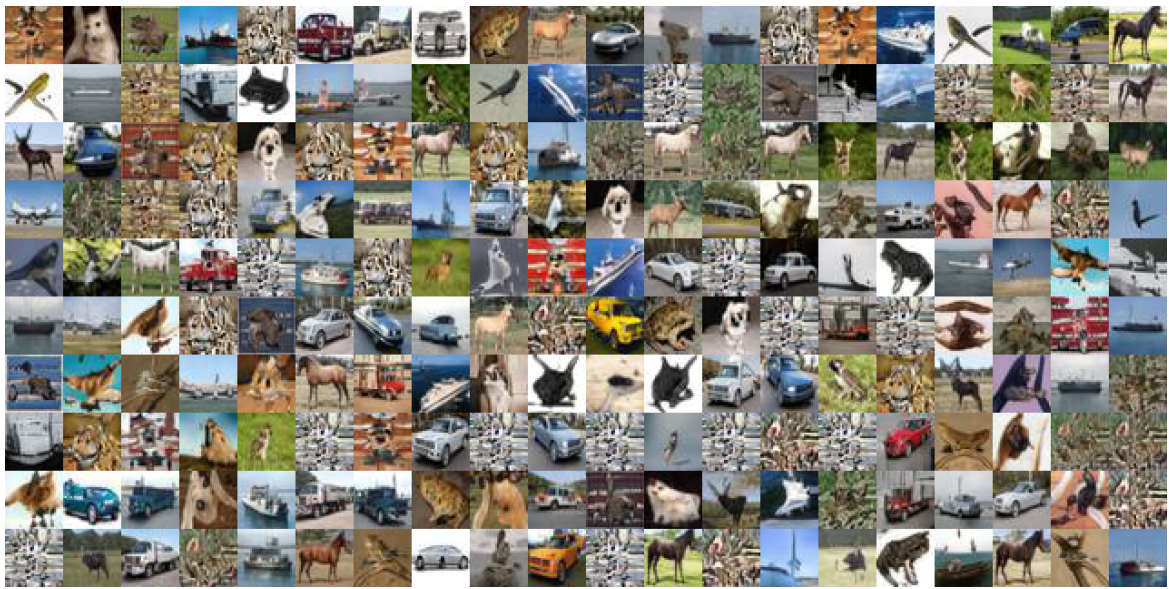}
  \end{center}
  % \vspace{-10pt}
  \caption{Mode collapse of RGM-KLD-D trained without the data fidelity term. Sampled images of RGM-KLD-D ($\lambda=\infty$) seem repetitive.} \label{fig:mode_collap}
  % \vspace{-12pt}
\end{figure}

\paragraph{Reducing mode collapse using data fidelity}
Lastly, we examine the influence of the data fidelity term in our MAP-based estimation.
To quantify the contribution of the fidelity term, we trained RGM-KLD-D by the loss function without the data fidelity loss (termed by RGM-KLD-D ($\lambda=\infty$)) in Section \ref{sec:ablation}, and we reached an FID score of 32.5 (See Table \ref{tab:ablation}). This result clearly motivates our objective. Moreover, we observe the mode collapse for RGM-KLD-D ($\lambda=\infty$), which is the one of common failure modes of GAN. As evidence, generated samples are presented in Figure \ref{fig:mode_collap}. Comparing samples generated by our RGM-KLD-D (see Figure \ref{fig:samples}) to Figure \ref{fig:mode_collap}, it is clear that images generated by RGM-KLD-D have higher diversity and better quality.
The results verify that it is beneficial to train our RGMs together with the data fidelity term.

\subsection{Comparison with existing models using various destruction} \label{appen:comparison_related}
Recently, several works introduce various degradation processes as an alternative to the diffusion process. \citet{rissanen2022generative} proposed an inverse heat dissipation model (IHDM) with a forward blurring process inspired by heat equation. 
Afterward, \citet{hoogeboom2022blurring} established a theoretical bridge between diffusion models and IHDM using Fourier transform. Based on this insight, they built a blurring diffusion model.
\citet{daras2022soft} proposed a general framework for learning the score function for any linear corruption process.
Moreover, Cold Diffusion \citep{bansal2022cold} proposed a new family of models using deterministic degradation processes.
Similarly, the proposed RGMs can leverage general linear degradation processes. Therefore, we compare the performance of RGMs with the aforementioned related works in Table \ref{tab:comparison_related}.
In comparison with our model itself, the change in the forward process brings FID improvement.
But compared to other models, we can observe how efficiently our proposed method produces high-quality images.

\begin{center}
\begin{table}[h]
    \centering
     \caption{Comparison with restoration-based models with various forward\\ processes. Sample quality on CIFAR10 is measured by FID score.} \label{tab:comparison_related}
     \vspace{-5pt}
    \scalebox{0.99}{
\begin{tabular}{cccc}
    \toprule
    Model & FID ($\downarrow$) & NFE \\
    \midrule
    Cold Diffusion (SR) \citep{bansal2022cold} &  152.76  & 3        \\
    Cold Diffusion (Blur) \citep{bansal2022cold} &   80.08  & 50   \\
    IHDM \citep{rissanen2022generative}& 18.96 & 200\\
    Soft Diffusion \citep{daras2022soft} & 3.86 & $\leq 100$ \\
    Soft Diffusion (Blur) \citep{daras2022soft} & 4.64 & $\leq 100$ \\
    Blurring Diffusion \citep{hoogeboom2022blurring}  &  3.17  & 1000           \\
    \midrule
    RGM-KLD-D\ \   &    3.04    & 4       \\
    RGM-KLD-SR & 2.47  & 7 \\
    \bottomrule
  \end{tabular}}
\end{table} 
\end{center}

\subsection{Additional Results on Inverse Problems}\label{appen:samples_inverse}
To quantify the performance of our RGM, we report signal-to-noise ratio (PSNR), which measures faithfulness to the ground-truth image. Also, as a perceptual metric, we include structural similarity index measure (SSIM) \citep{wang2004image} that quantifies the image.
Table \ref{tab:inverse} summarizes the PSNR and SSIM performances of colorization and super-resolution (SR) on CelebA-HQ and LSUN datasets.
Since the primary goal of SDEdit is to generate a realistic and faithful image in the absence of paired data, we did not make a quantitative comparison with SDEdit. But we include qualitative comparisons.

\begin{figure}
  \vspace{-5pt}
  \begin{center}
    \includegraphics[width=0.9\textwidth]{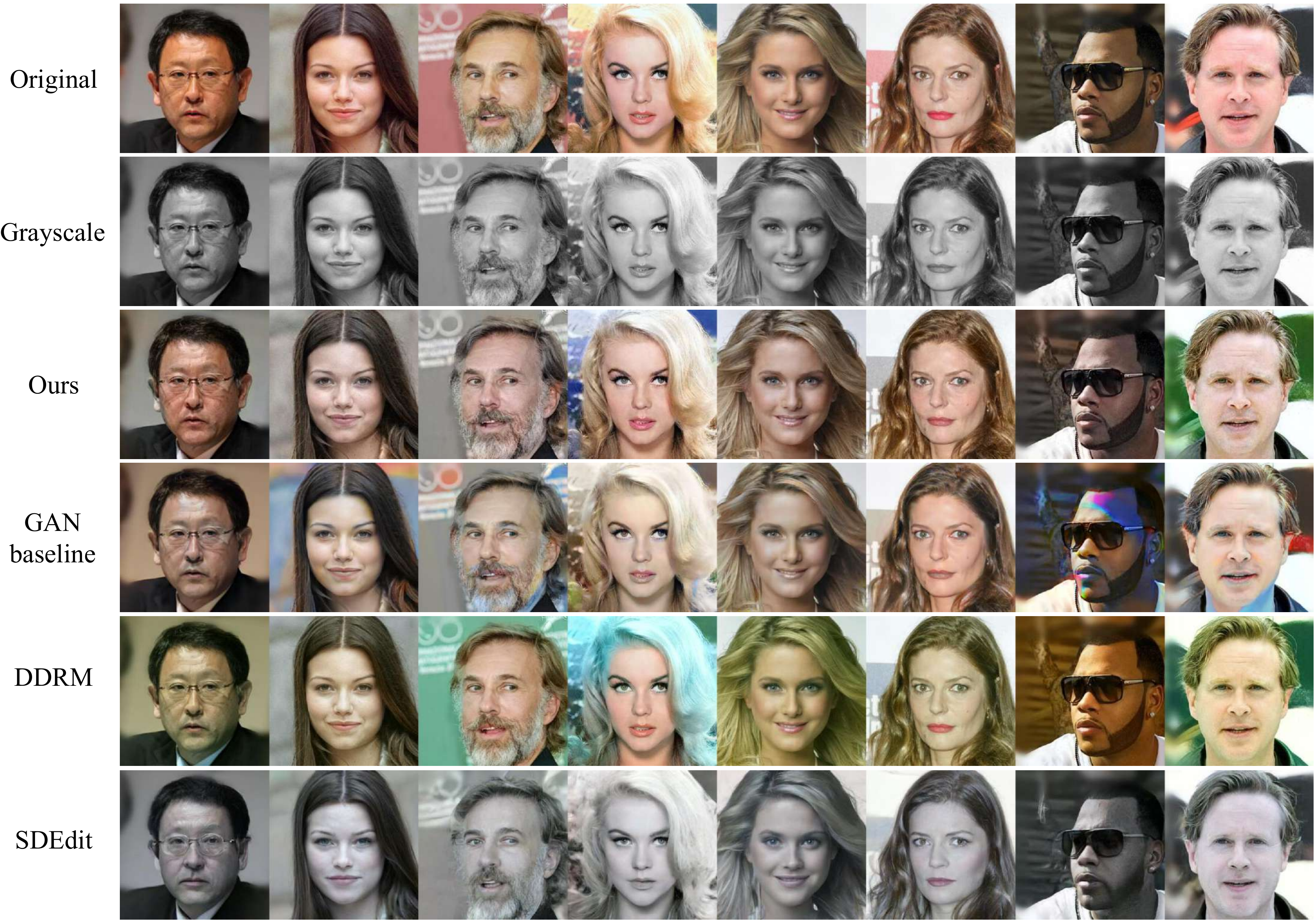}
  \end{center}
  \vspace{-10pt}
  \caption{\textbf{Colorization}. Qualitative comparison on CelebA-HQ.}
  \label{fig:color_celeba}
  \vspace{-12pt}
\end{figure}

\begin{figure}
  \vspace{-5pt}
  \begin{center}
    \includegraphics[width=0.9\textwidth]{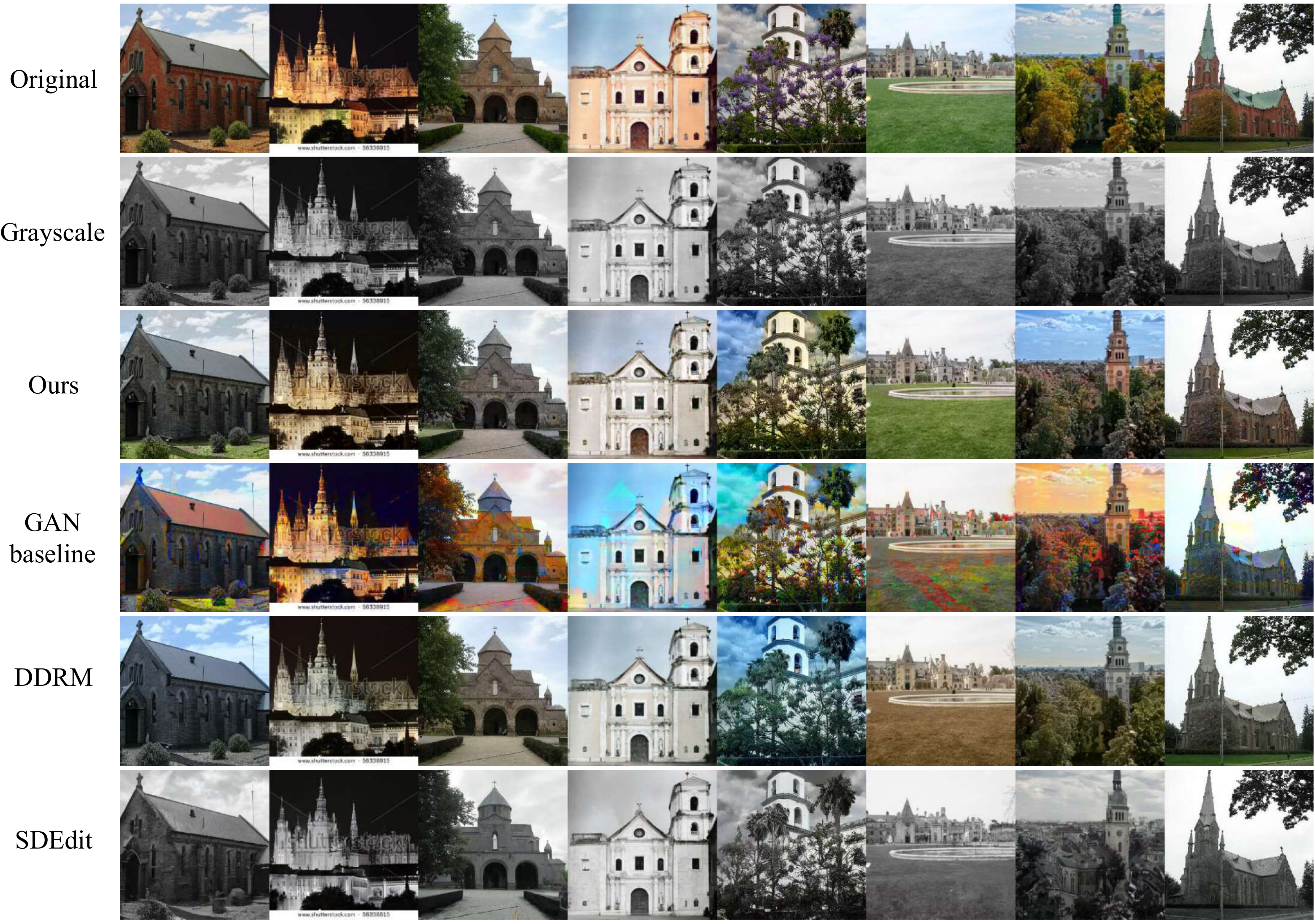}
  \end{center}
  \vspace{-10pt}
  \caption{\textbf{Colorization}. Qualitative comparison on LSUN church.}
    \label{fig:color_lsun}
  \vspace{-12pt}
\end{figure}

\begin{table}
    \centering
    \vspace{-5pt}
      \setlength\tabcolsep{3.0pt}
       \captionof{table}{Colorization and super-resolution results of different methods.} \label{tab:inverse}
       \vspace{-5pt}
       \scalebox{0.95}{
    \begin{tabular}{ccccccccccccc}
    \toprule
    Model & \multicolumn{4}{c}{Colorization}  & \multicolumn{8}{c}{Super-Resolution}\\   
    & \multicolumn{2}{c}{LSUN} & \multicolumn{2}{c}{CelebA-HQ} &\multicolumn{4}{c}{LSUN} & \multicolumn{4}{c}{CelebA-HQ}\\
     \cmidrule(lr){2-3}  \cmidrule(lr){4-5}  \cmidrule(lr){6-9}  \cmidrule(lr){10-13}
        &&&&&   \multicolumn{2}{c}{($\times 4$)} & \multicolumn{2}{c}{($\times 8$)} &   \multicolumn{2}{c}{($\times 4$)} & \multicolumn{2}{c}{($\times 8$)} \\ 
       &PSNR&SSIM&PSNR&SSIM&PSNR&SSIM&PSNR&SSIM &PSNR&SSIM&PSNR&SSIM\\
    \midrule
     RGM& 23.78 & 0.93& 25.57 & 0.93 & 22.74&0.65& 19.96& 0.48& 28.51&0.81& 24.86&0.70 \\
    DDRM  & 23.68& 0.94 &23.94& 0.93& 23.22&0.67&20.61&0.51&29.32&0.83&26.23&0.73  \\
    GAN baseline & 20.02&0.81&24.79&0.88 & 20.32&0.48&18.06&0.34&26.77&0.71&23.92&0.59  \\
    \bottomrule
  \end{tabular}}
    \vspace{-0pt}
\end{table}
\paragraph{Colorization}
The goal of image colorization is to restore a gray-scale image to a colorful image with RGB channels. We present more colorization results on CelebA-HQ and LSUN church in Figure \ref{fig:color_celeba} and \ref{fig:color_lsun}, respectively. 
Results reported in Table \ref{tab:inverse} show that our RGM achieves comparable and sometimes even better performance than baselines.
From the qualitative results, we can observe that our RGM is able to reconstruct more faithful and realistic images than other models.

\begin{figure}
  \vspace{-5pt}
  \begin{center}
    \includegraphics[width=0.85\textwidth]{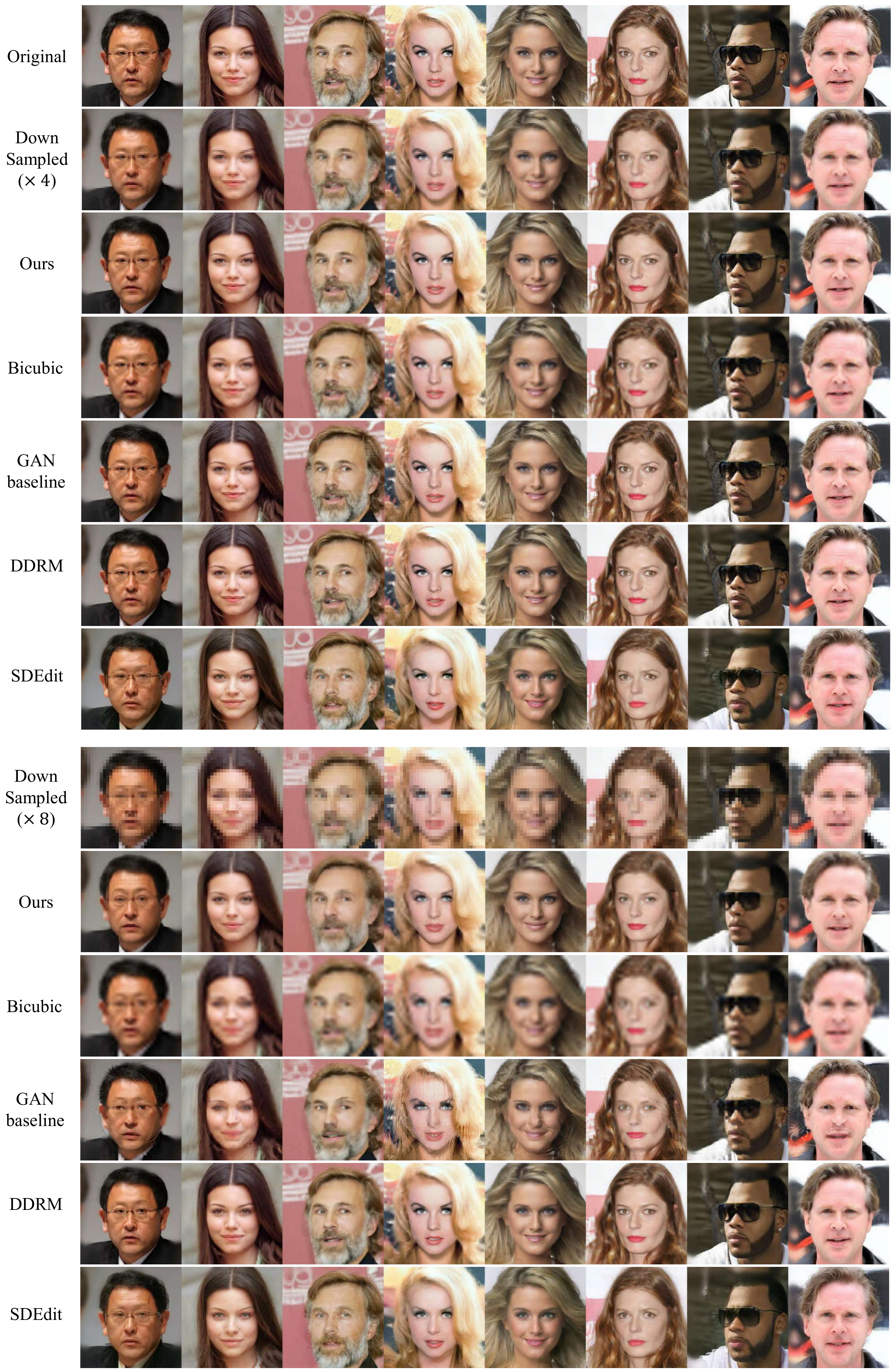}
  \end{center}
  \vspace{-10pt}
  \caption{\textbf{Super-resolution}. Qualitative comparison on CelebA-HQ.} \label{fig:sr_celeba}
  \vspace{-12pt}
\end{figure}

\begin{figure}
  \vspace{-5pt}
  \begin{center}
    \includegraphics[width=0.85\textwidth]{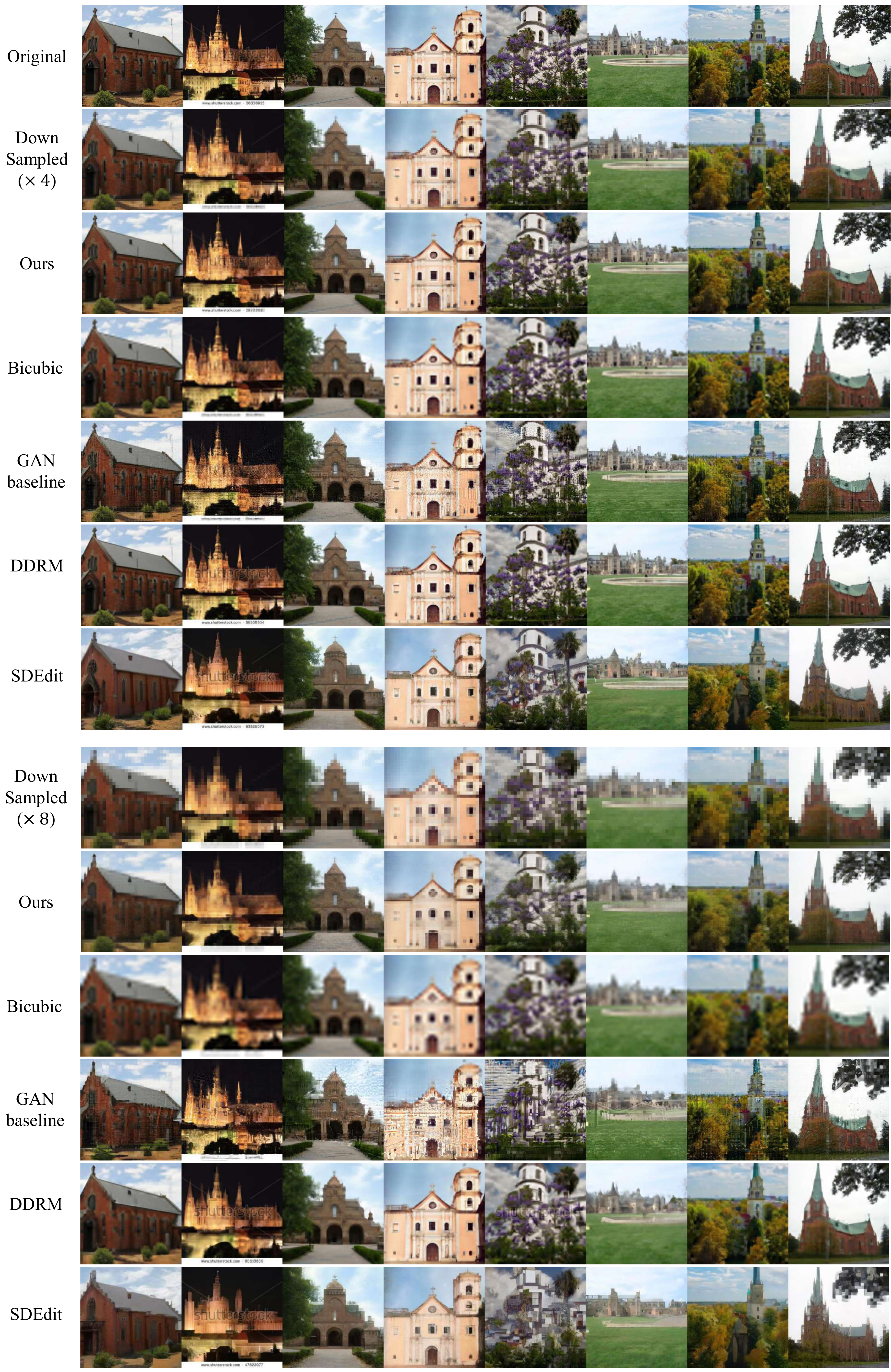}
  \end{center}
  \vspace{-10pt}
  \caption{\textbf{Super-resolution}. Qualitative comparison on LSUN church.} \label{fig:sr_lsun}
  \vspace{-12pt}
\end{figure}

\paragraph{Super-resolution}
Super-resolution aims at recovering high-resolution images corresponding to a given low-resolution image.
We consider downsampled images with two scale factors 4 and 8. We also compare SR results with bicubic interpolation.
Figure \ref{fig:sr_celeba} and \ref{fig:sr_lsun} present the qualitative comparisons.
Compared against bicubic upsampling, bicubic attains higher PSNR and SSIM values. However, we can observe from Figure \ref{fig:sr_celeba} and \ref{fig:sr_lsun} that bicubic interpolation results in blurry images, and RGM super-resolves more plausible images.
Also, visual differences between RGM and DDRM are qualitatively not large.

% \newpage

\begin{figure}
  % \vspace{-1pt}
  \begin{center}
  \includegraphics[width=0.6\textwidth]{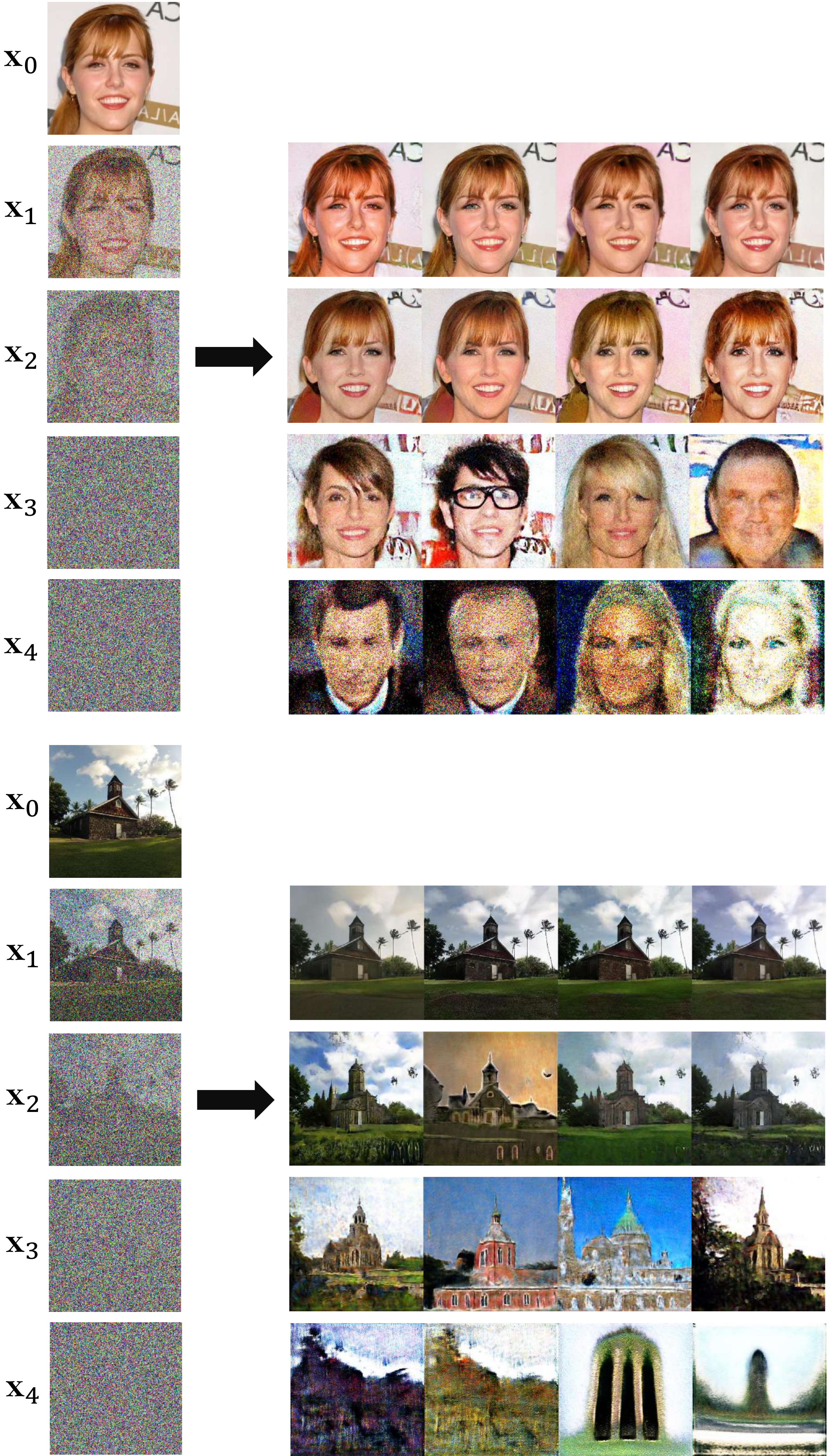}
  \end{center}
  \vspace{-5pt}
  \caption{Illustration of the effect of varying $\bz$ on CelebA-HQ (top) and LSUN (bottom). The images in the leftmost column depict the selected trajectory $\{\bx_k\}_{k=1}^4$ degraded from an image $\bx_0$. Each row on the right presents restored images of $\bx_t$ using four different random auxiliary values $\bz$. When the noise level is small, they generate almost identical images, which means that the restoration problem is almost well-posed. As the noise level increases, however, each degraded observation $\bx_k$ estimates diverse images depending on the $\bz$.
  In other words, the larger the noise, the more severe the ill-posedness, and the results validate that a much wider restoration is possible through the introduction of $\bz$.
  } \label{fig:z_appen}
  \vspace{-5pt}
\end{figure}

\subsection{Additional results of varying $\bz$} \label{appen:z_ablation}
We investigated the influence of the auxiliary variable $\bz$ in Section \ref{sec:ablation}. Here, we include more observations in Figure \ref{fig:z_appen}.  

\subsection{Additional Qualitative Results on Generation} \label{appen:samples}
\begin{figure}
  % \vspace{-5pt}
  \begin{center}
    \includegraphics[width=0.97\textwidth]{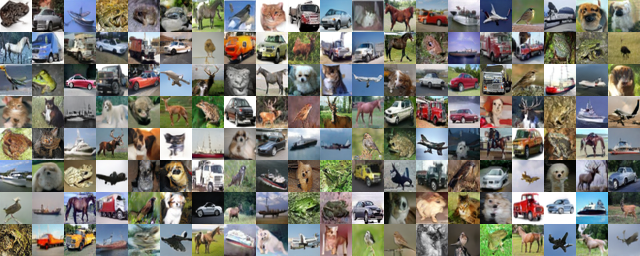}
  \end{center}
  % \vspace{-10pt}
  \caption{Generated samples of RGM-DSWD-D on CIFAR10.} \label{fig:cifar_dswd}
  % \vspace{-12pt}
\end{figure}

\begin{figure}
  % \vspace{-5pt}
  \begin{center}
    \includegraphics[width=0.97\textwidth]{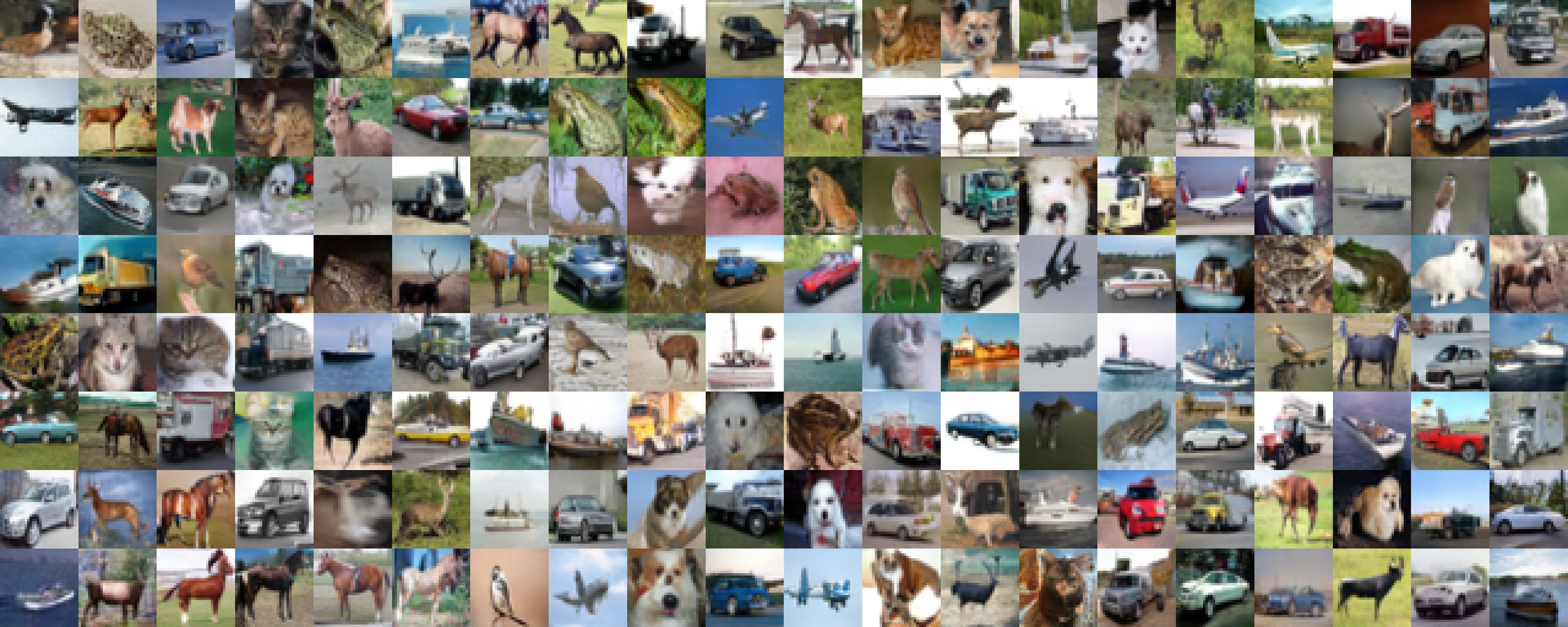}
  \end{center}
  % \vspace{-10pt}
  \caption{Generated samples of RGM-KLD-D on CIFAR10.} \label{fig:cifar_d}
  % \vspace{-12pt}
\end{figure}

\begin{figure}
  % \vspace{-5pt}
  \begin{center}
    \includegraphics[width=0.97\textwidth]{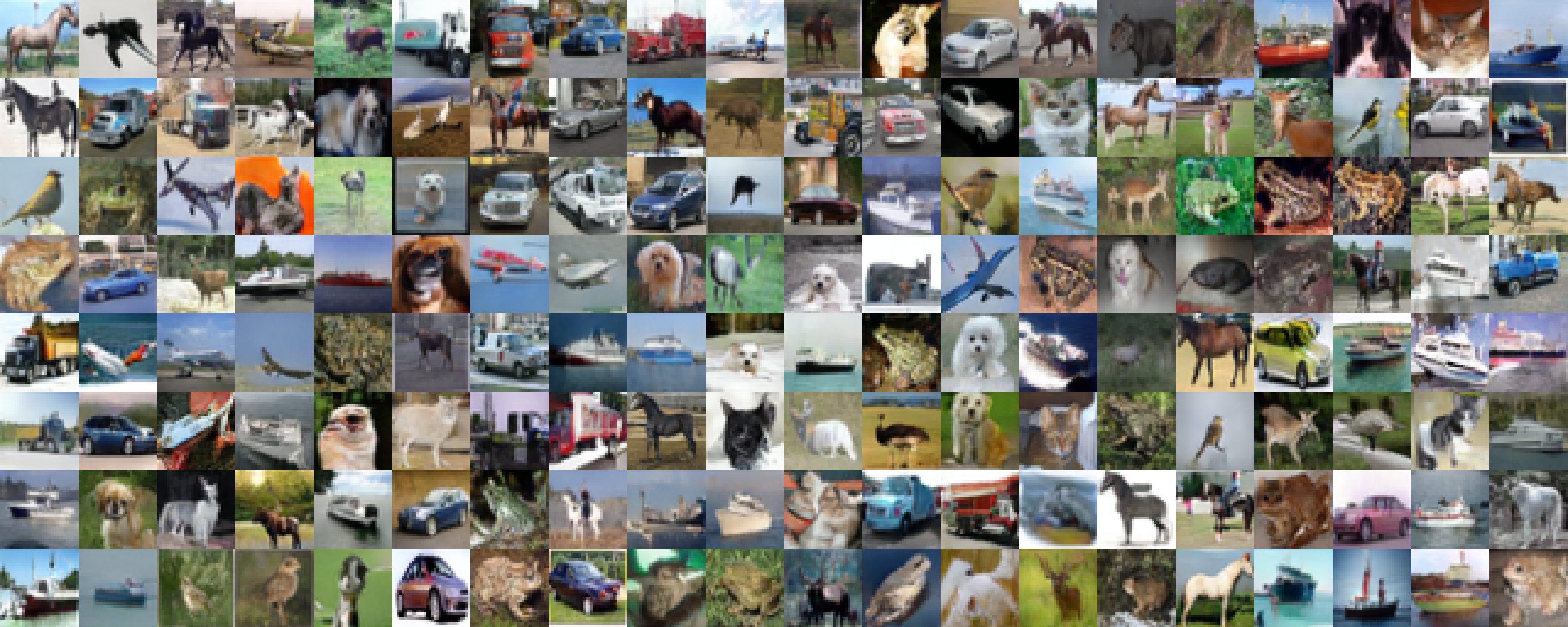}
  \end{center}
  % \vspace{-10pt}
  \caption{Generated samples of RGM-KLD-SR on CIFAR10.} \label{fig:cifar_sr}
  % \vspace{-12pt}
\end{figure}

\begin{figure}
  \vspace{-5pt}
  \begin{center}
    \includegraphics[width=0.9\textwidth]{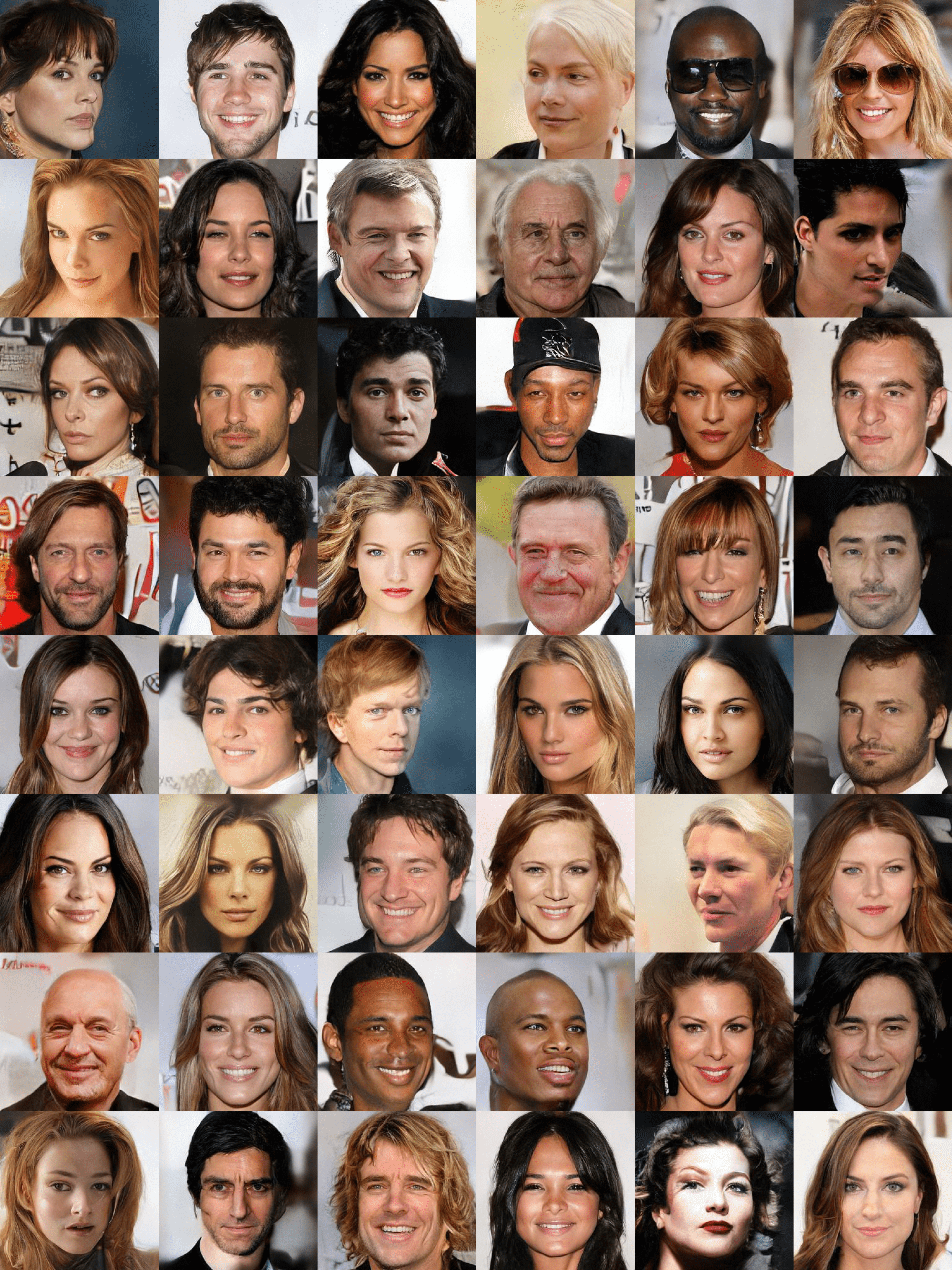}
  \end{center}
  \vspace{-10pt}
  \caption{Additional qualitative results of RGM-KLD-D trained on CelebA-HQ-256.} \label{fig:celeba}
  \vspace{-12pt}
\end{figure}

\begin{figure}
  \vspace{-5pt}
  \begin{center}
    \includegraphics[width=0.9\textwidth]{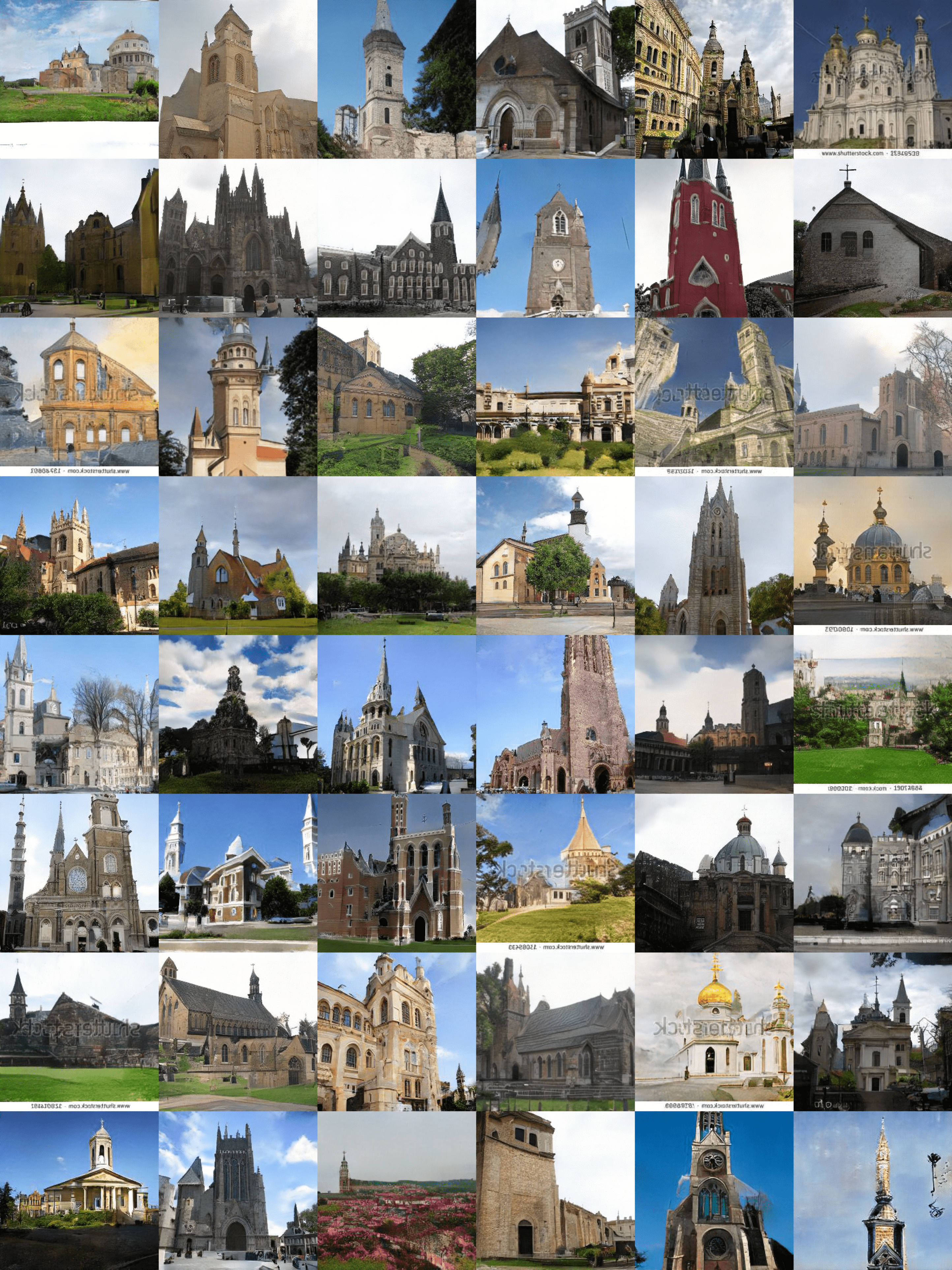}
  \end{center}
  \vspace{-10pt}
  \caption{More qualitative results of RGM-KLD-D trained on LSUN Church.} \label{fig:lsun}
  \vspace{-12pt}
\end{figure}

We present more generated image samples in Figures \ref{fig:cifar_dswd}, \ref{fig:cifar_d}, \ref{fig:cifar_sr}, \ref{fig:celeba}, and \ref{fig:lsun}.

\end{document}